\documentclass[lettersize,journal]{IEEEtran}
\usepackage{amsmath,amsfonts}
\usepackage{algorithmic}
\usepackage{algorithm}
\usepackage{array}
\usepackage[caption=false,font=normalsize,labelfont=sf,textfont=sf]{subfig}
\usepackage{textcomp}
\usepackage{stfloats}
\usepackage{url}
\usepackage{verbatim}
\usepackage{graphicx}
\usepackage{cite}
\usepackage{booktabs}
\usepackage{multirow}
\usepackage{subfig}
\usepackage{xcolor}
\usepackage{bm}
\begin{document}

\title{DecHW: Heterogeneous Decentralized Federated Learning\\  Exploiting Second-Order Information}

\author{
\IEEEauthorblockN{
Adnan Ahmad, 
Chiara Boldrini,
Lorenzo Valerio,
Andrea Passarella,
Marco Conti
}
\thanks{A.A. is with the School of Information Technology, Deakin University, Geelong, Australia. During the course of this work, A.A. was also a research fellow with the Institute for Informatics and Telematics (IIT), Italian National Research Council (CNR), Italy. 
C.B., L.V., A.P., and M.C. are with the Institute for Informatics and Telematics (IIT), Italian National Research Council (CNR), Italy.}
\thanks{Emails: {{adnan.a@deakin.edu.au}, \{chiara.boldrini, lorenzo.valerio, andrea.passarella, marco.conti\}@iit.cnr.it }}
\thanks{This work was partially supported by SoBigData.it. SoBigData.it receives funding from European Union – NextGenerationEU – National Recovery and Resilience Plan (Piano Nazionale di Ripresa e Resilienza, PNRR) – Project: “SoBigData.it – Strengthening the Italian RI for Social Mining and Big Data Analytics” – Prot. IR0000013 – Avviso n. 3264 del 28/12/2021. 
S. Sabella's, C. Boldrini's and M. Conti's work was partly funded by the PNRR - M4C2 - Investimento 1.3, Partenariato Esteso PE00000013 - ``FAIR'', A. Passarella's and L. Valerio's work was partially supported by the European Union - Next Generation EU under the Italian National Recovery and Resilience Plan (NRRP), Mission 4, Component 2, Investment 1.3, CUP B53C22003970001, partnership on ``Telecommunications of the Future'' (PE00000001 - program “RESTART”).}
}





\maketitle

\begin{abstract}
Decentralized Federated Learning (DFL) is a serverless collaborative machine learning paradigm where devices collaborate directly with neighbouring devices to exchange model information for learning a generalized model. However, variations in individual experiences and different levels of device interactions lead to data and model initialization heterogeneities across devices. Such heterogeneities leave variations in local model parameters across devices that leads to slower convergence. This paper tackles the data and model heterogeneity by explicitly addressing the parameter level varying evidential credence across local models. A novel aggregation approach is introduced that captures these parameter variations in local models and performs robust aggregation of neighbourhood local updates. Specifically, consensus weights are generated via approximation of second-order information of local models on their local datasets. These weights are utilized to scale neighbourhood updates before aggregating them into global neighbourhood representation. In extensive experiments with computer vision tasks, the proposed approach shows strong generalizability of local models at reduced communication costs.
\end{abstract}

\begin{IEEEkeywords}
Decentralized Federated Learning, Hessian matrix, Second-order Information
\end{IEEEkeywords}

\section{Introduction}
Federated Learning (FL) has become a cornerstone for collaborative machine learning (ML) across distributed clients, enabling ML models to be trained without requiring data to be centralized. Traditional FL architectures rely on a centralized server that coordinates the training process and aggregates client updates to create a global model \cite{McMahan2016CommunicationEfficientLO}.Despite its effectiveness in many use cases, the centralized approach has notable limitations. It is prone to system failures, as the server can become a single point of failure, and suffers from communication bottlenecks when scaled to millions of devices. Moreover, dependence on a central server can hinder direct client-to-client collaboration and pose practical challenges for scalability.


To tackle these challenges, Decentralized Federated Learning (DFL) has emerged as a new paradigm \cite{MartnezBeltrn2022DecentralizedFL}. DFL does not require a central server and enables devices to collaborate directly via a communication graph, often referred to as a communication network topology. Clients interact with a subset of their neighbors and exchange local updates to collectively build a generalized model. This serverless design promotes robustness and flexibility as all devices participate in a distributed social network that contributes to the development of a generalized model while mitigating the risks associated with centralized coordination~\cite{Palmieri2023}.

Despite its promising capabilities, DFL has inherited the challenges of data heterogeneity from traditional FL. Clients' local datasets often reflect differences in users' preferences, activities and contexts, leading to unbalanced sample distributions, different class representations and significant statistical variability. Furthermore, DFL introduces additional layers of complexity resulting from its decentralized nature. Unlike traditional FL, where a central server can enforce consistent model initialization and synchronization, DFL operates in an inherently heterogeneous and uncoordinated environment. Devices differ in computing capacities and network availability, leading to models with different initializations, training states and degrees of convergence for a given training task. These factors collectively contribute to both data and model heterogeneity, creating significant barriers to achieving a globally consistent and generalized model. The complex network topology further exacerbates these challenges, as devices can only communicate with their neighbors in the communication graph, making it difficult to achieve a robust generalized model in a communication-efficient manner.

A range of approaches have been proposed to tackle data and model heterogeneity \cite{ahmad2023federated, Karimireddy2019SCAFFOLDSC,ahmad_2023robust, Wang2020Federated,Ahmad2023RobustFL}. However, these methods are typically confined to server-coordinated FL and are not directly applicable to DFL. A few studies have also addressed data and model heterogeneity in DFL settings \cite{huang2024overcoming,Valerio2023CoordinationfreeDF}, but their performance in realistic DFL scenarios still leaves a significant gap. This raises the research question: 
\textit{How can a generalized model be achieved in a DFL scenario, where clients' models have different initializations and are trained on heterogeneous data?}
In this work, we investigate this question in a practical DFL scenario, characterized by clients connected via a communication network graph, heterogeneous data distributions, and diverse model initializations.

Heterogeneities in the model initialization and data distribution result in variations in the local model parameters, causing important parameters to vary across clients \cite{Wang2020Federated}. These variations in the model parameters lead to slower convergence when aggregating with conventional methods such as parameter averaging and must therefore be given special consideration before aggregation. Knowledge distillation (KD) based on shared synthetic data and parameter matching techniques  \cite{huang2024overcoming,Wang2020Federated} have also been proposed to tackle this problem. However, KD often depends heavily on shared synthetic or real datasets, which can be impractical to obtain in DFL settings.   Additionally, parameter matching approaches face significant challenges without a coordinating server to handle model expansion and correctly align neurons across diverse devices.
In this work, we explicitly account for quality variations in local models and introduce a novel approach to effectively aggregate models from neighboring clients. Our approach does not require data to be shared between devices and removes the necessity for a central coordinating server. \textcolor{black}{Unlike existing DFL approaches, which assign a uniform scalar weight to the entire model, typically proportional to the size of the local dataset, we address the parameter-level disparities across clients by introducing a weighted averaging strategy that leverages second-order information from local models. Rather than treating all parameters equally, our method captures quality variations in individual parameters and assigns parameter-level weights. These weights are then used during neighborhood model aggregation to scale each parameter according to its local importance. This approach ensures efficient aggregation, addressing the complexities introduced by heterogeneity in DFL environments.}


%
We summarize the key contributions of this work as follows:
\begin{itemize}
    \item We introduce Decentralized Hessian-Weighted Aggregation (DecHW), a novel aggregation method to effectively handle data and model initialization heterogeneity in DFL. 
    \item DecHW efficiently exploits the Hessian information in DFL by tracking its evolving trends across communication rounds and leverages these insights to develop a stable and robust aggregation strategy.

    \item We perform comprehensive experiments on image classification tasks to demonstrate the fast convergence of the proposed approach, highlighting its effectiveness in handling data and model initialization heterogeneity.
\end{itemize}

\section{Related Work}
Recently, DFL has emerged as a significant area of research. A number of studies have been proposed with focus on various areas ranging from industrial to medical applications. Roy et al. \cite{Roy2019BrainTorrentAP} proposed a DFL framework for medical applications, enabling multiple parties to collaboratively train ML models on their local datasets. Lalitha et al. The approach proposed by Lalitha et al. \cite{Lalitha2019DecentralizedBL} leverages Bayesian principles to reduce the Kullback-Leibler (KL) divergence between local and aggregated peer models. Savazzi et al. \cite{Savazzi2019FederatedLW} introduce a consensus federated algorithm (CFA) that extends the Federated Averaging model to decentralized settings, especially suited for industrial and IoT use cases. Sun et al. \cite{Sun2021DecentralizedFA} propose a Federated Decentralized Average method, which incorporates momentum to reduce drift caused by multiple stochastic gradient descent (SGD) updates and employs a quantization scheme to improve communication efficiency.


Another approach \cite{Valerio2023CoordinationfreeDF} aims to address data and initialization heterogeneity by enabling each device to consider the degree of dissimilarity when aggregating its local model with those from neighboring devices. However, this method still relies on parameter averaging to combine neighboring models, which limits its effectiveness. A more recent work \cite{huang2024overcoming} introduces an alternative to parameter averaging through the use of Knowledge Distillation applied on a shared synthetic dataset. This approach first accumulates a balanced synthetic dataset before the DFL process. Instead of sharing local model parameters, neighboring devices exchange output logits on synthetic data, and KD is applied to transfer knowledge from these logits. While this method improves communication efficiency, the creation and distribution of synthetic data remain impractical in DFL.

Our solution differs in that we consider the sensitivity of the local models at the parameter level and weigh each parameter according to its importance for the local data distribution. Specifically, we utilize the diagonal of the Hessian as a curvature-aware weighting mechanism during model aggregation, enabling each device to contribute proportionally based on parameter stability. Unlike traditional second-order optimization techniques such as AdaGrad \cite{JMLR:v12:duchi11a}, which use curvature information to adapt local learning rates, our method employs second-order information directly for aggregation in decentralized settings. 
Our method does not require shared datasets (synthetic or real), operates without a coordinating server, and introduces minimal computational and communication overhead. Moreover, it consistently outperforms other approaches in terms of speed and accuracy across the tested datasets.

\vspace{-5pt}
\section{Problem setup}
Consider a static graph $\mathcal{G}(\mathcal{N},\mathcal{E})$, where $\mathcal{N}$ denotes the devices (nodes) and $\mathcal{E}$ the communication links connecting them. Without loss of generality, the focus is on a static topology in which the links remain unchanged over time. Each device $i \in \mathcal{N}$ holds a local dataset $D_i = \{(x_k, y_k)\}_{k=1}^{|D_i|}$ with size $|D_i|$. Here, $x_k \in \mathcal{X}_i$ represents the input feature space for device $i$, and $y_k \in \mathcal{Y}_i$ denotes the corresponding label space. The union of local feature spaces, $\bigcup_{i \in \mathcal{N}} \mathcal{X}_i$, forms the global input space $\mathcal{X}$. Each device $i \in \mathcal{N}$ maps its local feature space $\mathcal{X}_i$ to its corresponding label space $\mathcal{Y}_i$ through a parametric model $h(\cdot; w_i)$, where $w_i$ represents the model parameters. Nodes aim to achieve a shared global model $w^*$ to map $\mathcal{X}$ to $\mathcal{Y}$.

In DFL, each node $i\in \mathcal{N}$ aims to minimize its local empirical risk:

\vspace{-10pt}
\begin{equation} 
    w_i^* = \arg\min_{w_i} \mathcal{L}_i(w_i) = \frac{1}{|D_i|} \sum_{(x_k, y_k) \in D_i} \ell(h(x_k; w_i), y_k)
\end{equation}
where $\mathcal{L}_i(\cdot)$ represents the local loss function at node $i$ and $l(\cdot,\cdot)$ represents a loss function suitable for the learning task at hand, e.g., a typical choice for classification tasks is represented by Cross-Entropy loss.

In DFL, the absence of a coordinating server requires devices to communicate directly with one another. Communication is governed by the graph $\mathcal{G}(\mathcal{N}, \mathcal{E})$, which specifies the connections between devices. Each device $i$ can exchange information only with its neighbors, defined by the neighborhood set $\mathcal{N}_i$. If a direct link is absent, devices cannot collaborate directly and must rely on intermediaries to propagate information. Thus, to achieve a shared global model, node $i$ exchanges local parameters with its neighbors $\mathcal{N}_i$. The aggregation of the local models can be represented as:
\begin{equation}
    w_i^{(t)} = \psi(w_i^{(t-1)},\{w_j^{(t-1)}\}_{j\in\mathcal{N}_i})
\end{equation}
where $\psi$ represents a generic aggregation function, which combines the models (the local one from $i$ and the ones coming from $i$'s neighbors), and  $w_i^{t}$ is the updated model parameter for node $i$ at communication round $t$. Similar to server-coordinated FL, weighted averaging is commonly used as $\psi$ to aggregate local updates of neighbors, and it is given by:
%
%
\begin{equation} \label{eq:fedavg_aggr}
    w_i^{(t)} = \sum_{\forall j\in \mathcal{N}_i \cup i}  \tau_j w_j^{(t-1)}
\end{equation}
where $\tau_j=\frac{|D_j|}{\sum_{\forall k\in \mathcal{N}_i \cup i}|D_k|}$ weights the parameters $w_j$ proportionally to the amount of data the node $j$ holds w.r.t. those present in the neighborhood of node $i$.

\begin{figure}
    \centering
    \includegraphics[width=0.92\linewidth]{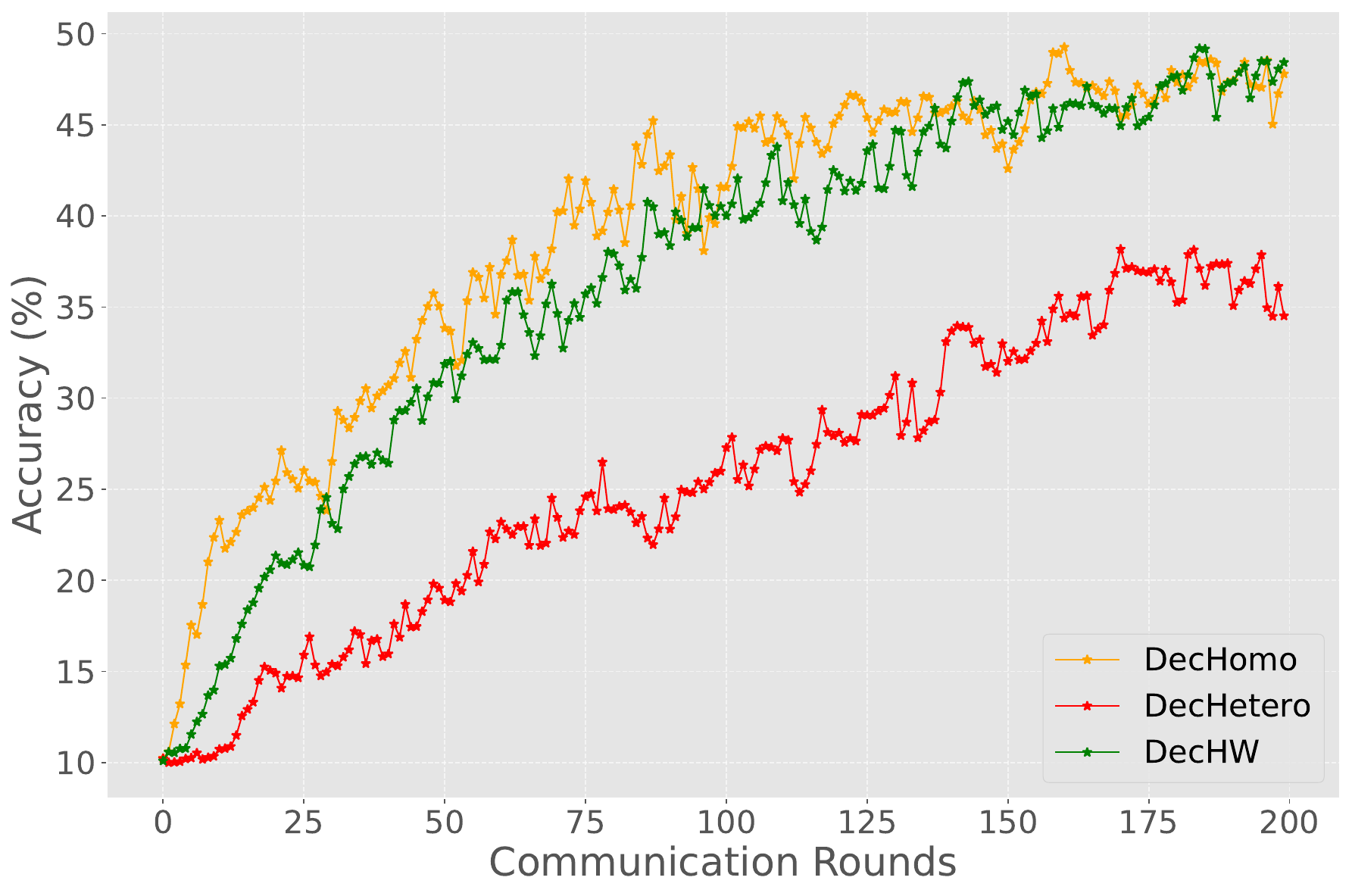}
    \caption{Average accuracy across devices on the CIFAR10 dataset in DFL settings (details in Sec.~\ref{sec:exp_setup}). \textit{DecHomo}: Parameter-wise averaging with uniform model initialization. \textit{DecHetero}: Parameter-wise averaging with diverse model initializations. \textit{DecHW}: Proposed method with diverse model initializations.}
    \label{fig:acc_homo}
\end{figure}

The aggregation approach described in Equation \ref{eq:fedavg_aggr} relies on proper neuron alignment to ensure robust aggregation. However, this alignment is often disrupted by model initialization and data heterogeneity, which are typical of DFL settings. Due to differences in model initialization, data distribution, training states, and levels of convergence, the local model parameters are often misaligned across devices. Aggregating such misaligned parameters with a simple weighted average can result in degraded performance \cite{Wang2020Federated}. Figure~\ref{fig:acc_homo} illustrates this problem by comparing two initialization strategies in a DFL setup. When all models are initialized homogeneously, the standard parameter-wise averaging converges efficiently. However, when models are initialized heterogeneously, the performance significantly degrades, requiring more communication rounds to achieve convergence. In addition, these challenges are further complicated by data heterogeneity and the different number of neighbors for each device.
This highlights the detrimental effects of heterogeneity on aggregation efficiency, as misalignment of parameters increases inconsistency between local models and hinders progress towards a generalized global model.

\section{Proposed Methodology}
\begin{figure}
    \centering
    \includegraphics[width=0.98\linewidth]{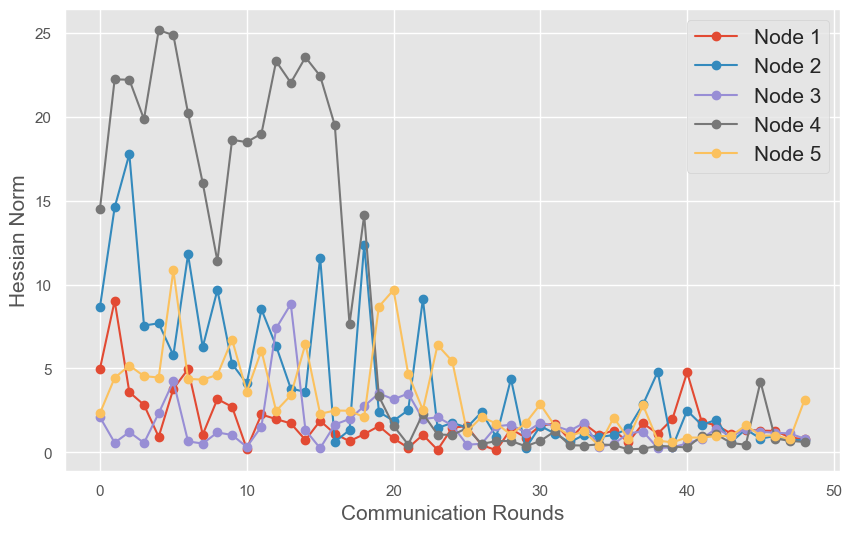}
    \caption{The norm of the Hessian diagonal was observed for a few nodes during training on the MNIST dataset, as outlined in Sec. \ref{sec:exp_setup}. As training progresses, the norms gradually decrease towards zero.}
    \label{fig:hess_norm}
\end{figure}

While existing aggregation approaches using Equation (\ref{eq:fedavg_aggr}) scale model updates according to the size of each client’s local dataset, they assign equal weight to all intra-model parameters, regardless of their importance or sensitivity. This can lead to poor generalization, especially in scenarios where the data distributions across devices are heterogeneous. To overcome these limitations, we introduce a weighted aggregation strategy that incorporates the sensitivity of each local model parameter, accounting for its importance within the local data distribution. To capture the importance of local model parameters, we aim to exploit the second-order information of the local models. Specifically, we incorporate the Hessian matrix to evaluate the sensitivity of each parameter in the local data distribution. \textcolor{black}{Intuitively, the Hessian of the loss with respect to a parameter reflects the curvature of the loss surface in that dimension, indicating the parameter’s sensitivity to small perturbations. A large Hessian value implies that even a small change in the parameter would substantially affect the loss, indicating high importance for the local data distribution. On the other hand, a small Hessian value suggests low sensitivity and thus lower importance. Leveraging this property enables us to assign greater weight to parameters that are more critical for each client, leading to more informed and effective aggregation under heterogeneous conditions.}

\subsection{Hessian Matrix}
The Hessian, a square matrix, consists of the second-order partial derivatives of the loss function. The Hessian matrix, \( H_i \), of the local loss function, \( l_i(w_i) \), at device \( i \) captures the curvature of the loss function with respect to the model parameters \( w_i \) \cite{NEURIPS2018_18bb68e2}. The Hessian provides information on how parameter changes affect the local loss. Specifically, the diagonal elements of the Hessian matrix, \( H_{i,\text{diag}} \), represent the sensitivity of the loss to changes in each parameter:
\[
H_{i,\text{diag}} = \left[ \frac{\partial^2 l_i(w_i)}{\partial w_{i,1}^2}, \frac{\partial^2 l_i(w_i)}{\partial w_{i,2}^2}, \ldots, \frac{\partial^2 l_i(w_i)}{\partial w_{i,m}^2} \right]
\]
Here, \( m \) denotes the number of parameters in \( w_i \). However, a major limitation of using the Hessian is its high computational cost, which makes it impractical for optimization problems invloving large number of parameters \cite{LeCun2012EfficientB}. To overcome this, the Gauss-Newton optimization method \cite{Becker1989ImprovingTC,Schraudolph2001FastCM} is commonly employed as a more efficient alternative to the Newton method given by $   H_{i,\text{diag}} = \bm{J}(l_i(w_i))^T\bm{J}(l_i(w_i))$, 
where, $\bm{J}$ represents the Jacobian matrix of the gradient of the loss with respect to parameters $w_i$.

In ML, particularly in optimization algorithms such as Newton's method \cite{Nocedal2018NumericalO} and its variants \cite{Shamir2013CommunicationEfficientDO,Islamov2021DistributedSO}, it is common to adjust the learning rate using the inverse of the Hessian, which helps prevent destabilizing updates to sensitive parameters. In contrast, in FL, if a parameter in the local model is less sensitive, it could indicate that it has already reached a flat region from the perspective of the local data during the current training round. This flatness often results from overfitting to the limited dataset available at each device.
Large Hessian diagonal entries indicate parameters that have not converged and are still important for minimizing the loss, as they correspond to directions in the parameter space where changes significantly affect the loss. On the other hand, small or near-zero Hessian diagonal entries suggest that the parameters may be in flat regions of the loss surface, where changes to these parameters have little effect on the loss, potentially indicating overfitting. In light of these considerations, the rationale of our method is as follows: \emph{we give more weight to parameters with high sensitivity in the aggregation process to ensure that important parameters have a stronger influence on the generalized model}.

\subsection{Decentralized Hessian Weighted Aggregation}
To ensure that parameters with high sensitivity have a stronger influence on the global model update, we propose to consider the contribution of each local model \( w_i \) according to the normalized Hessian diagonal entries: 
\begin{equation}
\hat{H}_{i,\text{diag}} = \frac{H_{i,\text{diag}}}{\|H_{i,\text{diag}}\|_2},
\end{equation}
where \( \|H_{i,\text{diag}}\|_2 \) is the \( \ell_2 \)-norm of the diagonal of the Hessian matrix. The normalization of the Hessian diagonal ensures that all parameters are scaled according to their relative importance. 
\textcolor{black}{As shown in Figure~\ref{fig:hess_norm}, model parameters are highly sensitive to changes in the loss function during the early phase of training, while in later phases, overfitting causes the Hessian diagonal terms to approach zero, reflecting flat regions created by intra-client local minima. Both the excessive sensitivity in the early phase and the reduced sensitivity in the later phase can compromise aggregation and distort the relative importance of parameters. Properly accounting for these changes is essential to preserve meaningful parameter information and reduce short-term fluctuations in the Hessian values. To mitigate these variations, we leverage historical data on local Hessian diagonals and accumulate recent updates scaled by a factor $\beta$, which can be expressed as:}

%
\begin{equation}\label{eq:hessian_update}
\hat{H}_{i,\text{diag}}^{(t-1)} = \hat{H}_{i,\text{diag}}^{(t-2)} + \beta \frac{H_{i,\text{diag}}^{(t-1)}}{\|H_{i,\text{diag}}^{(t-1)}\|_2}
\end{equation}
where $\hat{H}_{i,\text{diag}}^{(t-1)}$ is the updated Hessian diagonal of device $i$ and $\beta\in[0,1]$ is a hyperparameter that controls the sensitivity of the approximate $\hat{H}_{i,diag}$ to new updates. \textcolor{black}{Appendix D provides a detailed evaluation of the impact of this Hessian accumulation.}


At end of the previous communication round $t-1$, each device $i$ exchanged both it's model parameters $w_i^{(t-1)}$ and Hessian diagonals $\hat{H}_{i,\text{diag}}^{(t-1)}$. For each parameter \( n \in \{1, 2, \ldots, m\} \), the weight \( p_{i,n} \) for device \( i \) is computed based on the normalized Hessian diagonal:
\vspace{-2pt}
\begin{equation}
    p_{i,n}^{(t)} = \frac{\hat{H}_{i,n}^{(t-1)}}{\sum_{j=1}^{N} \hat{H}_{j,n}^{(t-1)}}
\end{equation}
where $N=|\mathcal{N}_i|+1$. This weight \( p_{i,n} \) ensures that parameters with higher sensitivity have a larger influence on the generalized model update. The generalized model parameters \( w_{i,n}^{(t)} \) are updated by taking a weighted average of the local parameters, where the weights are given by the Hessian-based sensitivity values:
\vspace{-5pt}
\begin{equation}
w_{i,n}^{(t)} = \sum_{j=1}^{N} p_{j,n}^{(t)} w_{j,n}^{(t-1)},
\end{equation}
\( w_{j,n}^{(t)} \) is the \( n \)-th parameter of client \( j \)'s model at round \( t \), and \( w_{i,n}^{(t)} \) is the aggregated \( n \)-th parameter in the generalized model at node $i$.

However, certain parameters may remain largely insensitive to the data distribution, resulting in minimal information being captured in their corresponding Hessian diagonal entries (i.e., for parameter $n$, $p_{j,n}^{(t)} \sim 0$ for all $j$). To tackle this problem, we aggregate these insensitive parameters by scaling them according to the amount of data available at each device (i.e., we resort to standard averaging). The rationale is that models trained on larger datasets are more likely to generalize better, making their contributions to the global model more reliable. This ensures that even parameters with minimal sensitivity are aggregated in a meaningful way. Therefore, we incorporate a hybrid aggregation approach given by:
\begin{equation} \label{eq:hybrid}
w_{i,n}^{(t)} = 
\begin{cases}
\sum_{j=1}^{N} p_{j,n}^{(t)} w_{j,n}^{(t-1)} & \text{if } \sum_{j=1}^{N} p_{j,n}^{(t)} \neq 0 \\
 \sum_{j=1}^{N} \tau_j w_j^{(t-1)} & \text{otherwise}
\end{cases}
\end{equation}

%
%
We refer to our proposed Hessian-weighted learning strategy as DecHW.  \textcolor{black}{The pseudocode is presented in Appendix \ref{app:pseudo}, and the implementation is publicly available at \footnote{\url{https://anonymous.4open.science/r/DecHW-6B4E/}}.} 
In DecHW, devices update their models using the equation above during each communication round. As demonstrated in the next section, the second-order information leveraged by the Hessian-weighted aggregation not only substantially accelerates the transient phase of the decentralized learning process but also delivers superior performance at convergence. 



\section{Experiments}
\subsection{Experimental setup} \label{sec:exp_setup}
Following \cite{Valerio2023CoordinationfreeDF}, we adopt a static network graph topology to simulate the DFL scenario. Specifically, an Erd\H{o}s–R\'enyi graph \cite{Erdos1984OnTE} with 50 nodes is employed, where each edge is created independently with a probability of \( p = 0.2 \). This setup implies that almost surely each device is connected to at least one neighboring device (being $p$ above the critical threshold $\ln(|\mathcal{N}|)/|\mathcal{N}|$) but at the same time not densely connected (the density is exactly equal to $p$).

\begin{table*}[t]
  \centering
    \caption{ Mean $\pm$ Std of test Accuracy in last 10 communication rounds when CNN is trained on MNIST and Fashion dataset.}
  \resizebox{\textwidth}{!}{%
    \begin{tabular}{cccccccccccc}
      \hline
      Method & \multicolumn{3}{c}{MNIST} & \phantom{a} & \multicolumn{3}{c}{Fashion} & \phantom{a} & \multicolumn{3}{c}{CIFAR10} \\
      \cline{2-4} \cline{6-8} \cline{10-12}
      & dir(1) & dir(0.5) & dir(0.2) && dir(1) & dir(0.5) & dir(0.2) && dir(1) & dir(0.5) & dir(0.2)  \\ \hline
       DecHetero        & $90.93 \pm 0.005$ & $86.76 \pm 0.011$ & $88.14 \pm 0.013$ && $81.47 \pm 0.007$ & $81.24 \pm 0.010$ & $76.39 \pm 0.013$ && $52.81 \pm 0.008$ & $56.67 \pm 0.023$ & $51.28 \pm 0.027$ \\
       CFA              & $91.49 \pm 0.013$ & $88.10 \pm 0.018$ & $90.44 \pm 0.012$ && $81.82 \pm 0.007$ & $80.80 \pm 0.011$ & $76.82 \pm 0.024$ && $52.55 \pm 0.016$ & $54.29 \pm 0.035$ & $47.86 \pm 0.052$ \\
       DecDiff          & $89.60 \pm 0.024$ & $84.94 \pm 0.030$ & $88.40 \pm 0.013$ && $81.98 \pm 0.006$ & $81.73 \pm 0.007$ & $77.24 \pm 0.018$ && $49.24 \pm 0.030$ & $52.09 \pm 0.026$ & $51.83 \pm 0.014$ \\
       DecDiff (VT)     & $93.34 \pm 0.010$ & $90.48 \pm 0.009$ & $92.43 \pm 0.010$ && $82.59 \pm 0.004$ & $81.90 \pm 0.008$ & $78.47 \pm 0.010$ && $50.18 \pm 0.019$ & $53.54 \pm 0.025$ & $53.50 \pm 0.011$ \\

       DecHW            & $\bm{93.50 \pm 0.005}$ & $\bm{91.04 \pm 0.004}$ & $\bm{93.89 \pm 0.009}$ && $\bm{83.27 \pm 0.006}$ & $\bm{82.91 \pm 0.004}$ & $\bm{80.36 \pm 0.011}$ && $\bm{55.29 \pm 0.019}$ & $\bm{57.16 \pm 0.018}$ & $\bm{54.37 \pm 0.020}$\\

       \bottomrule
      
    \end{tabular}%
  }
  \label{tab:accuracy}
\end{table*}

\begin{table*}[t]
  \centering
    \caption{ Average time (in terms of communication rounds) required to achieve 50\%, 75\%, 90\%, and 95\% of best accuracy achieved on each dataset.}
  \resizebox{\textwidth}{!}{%
    \tiny
    \begin{tabular}{cccccccccccccccc}
      \toprule
      Dir ($\alpha$) & Method & \multicolumn{4}{c}{MNIST} & \phantom{a} & \multicolumn{4}{c}{Fashion} & \phantom{a} & \multicolumn{4}{c}{CIFAR10} \\
      \cmidrule{3-6} \cmidrule{8-11} \cmidrule{13-16}
      & & $50 \%$ & $75 \%$ & $90 \%$ & $95 \%$ && $50 \%$ & $75 \%$ & $90 \%$ & $95 \%$ && $50 \%$ & $75 \%$ & $90 \%$ & $95 \%$  \\ \midrule
       \multirow{6}{*}{$\alpha = 1$} & DecHetero      & 321 & 405 & 517 & 749 && 6 & 14 & 138 & 377 && 101 & 319 & 667 & 945\\
            & CFA              & 256 & 324 & 417 & 646 && 5 & 14 & 111 & 308 && 81 & 234 & 555 & 823\\
            & DecDiff          & 150 & 178 & 382 & 736 && 1 & 13 & 118 & 311 && 98 & 278 & 804 & -\\
            & DecDiff (VT)     & 168 & 189 & 300 & 455 && 2 & 12 & 84 & 255 && 100 & 279 & 703 & -\\
            & DecHW            & \textbf{10 }& \textbf{21 }&\textbf{ 108 }& \textbf{267 }&& 1 & \textbf{10 }& \textbf{41} & \textbf{159} &&\textbf{ 53} &\textbf{ 124} & \textbf{337} & \textbf{591}\\
       \midrule

       \multirow{6}{*}{$\alpha = 0.5$} & DecHetero       & 247 & 335 & 629 & 939 && 6 & 18 & 142 & 305 && 117 & 269 & 503 & 650 \\
            & CFA              & 215 & 245 & 485 & 775 && 5 & 13 & 92 & 233 && 125 & 241 & 524 & 874\\
            & DecDiff          & 44 & 116 & 589 & - && 8 & 16 & 104 & 280 && 121 & 283 & 566 & - \\
            & DecDiff (VT)     & 45 & 105 & 351 & 594 && 8 & 17 & 97 & 273 && 113 & 274 & 533 & -\\
            & DecHW            & \textbf{11} &\textbf{ 29} &\textbf{ 163} & \textbf{333} && \textbf{4} & \textbf{11} & \textbf{59} & \textbf{172} &&\textbf{ 44} & \textbf{108} &\textbf{ 264} & \textbf{334}\\
       \midrule
       \multirow{6}{*}{$\alpha = 0.2$} & DecHetero        & 189 & 379 & 909 & - && 8 & 46 & 267 & 673 && 218 & 557 & 810 & 998\\
            & CFA              & 77 & 249 & 580 & 913 && \textbf{6} & 30 & 238 & 516 && 170 & 356 & 608 & 817\\
            & DecDiff          & 37 & 280 & 953 & - && 11 & 38 & 227 & 498 && 156 & 336 & 609 & 729\\
            & DecDiff (VT)     & 36 & 145 & 376 & 544 && 12 & 30 & \textbf{127} & 388 && 170 & 332 & 539 & 651\\
            & DecHW            &\textbf{ 14 }& \textbf{46} &\textbf{ 148 }& \textbf{291 }&& 9 & \textbf{19} & \textbf{127} & \textbf{343} && \textbf{59} &\textbf{ 150} &\textbf{ 303} & \textbf{517}\\
       \bottomrule
    \end{tabular}%
  }
  \label{tab:communication_effic}
\end{table*}

\subsubsection{Datasets}
We consider image classification tasks in our experiments, utilizing the MNIST, Fashion-MNIST, and CIFAR-10 datasets. MNIST consists of 60,000 training examples and 10,000 test examples, with each example being a grayscale image of size $28 \times 28$ representing a single handwritten digit from 0 to 9. The dataset reflects diverse handwriting styles, making it particularly suitable for evaluating methods in DFL. Fashion-MNIST, a dataset of ten different fashion items, shares the same structure as MNIST in terms of size and number of samples, but provides a more challenging alternative. CIFAR-10 consists of 60,000 color images divided into 50,000 training examples and 10,000 test examples, with each image being of size $32 \times 32$ and belonging to one of ten classes. CIFAR-10 introduces additional complexity with its larger image size, more channels, and varied image content.

\subsubsection{Data allocation}
Following standard practices~\cite{Wang2020Federated}, we use a Dirichlet distribution $Dir(\alpha)$ with $\alpha = \{0.2, 0.5, 1\}$ to distribute data among devices. A lower value of $\alpha$ increases the data heterogeneity, creating more diverse data distributions across devices, while a higher value of $\alpha$ reduces heterogeneity, resulting in more balanced data distributions. More information about data distributions is provided in appendix \ref{app:data_dist}.
\subsubsection{Model architecture and configuration}
We consider Convolutional Neural Networks (CNNs) for all three datasets. For MNIST, we use two convolutional layers (10 and 20 filters, \(5 \times 5\) kernels), max pooling, and two fully connected layers (320, 50 units). For Fashion-MNIST, we use two convolutional layers (32, 64 filters, \(3 \times 3\) kernels), max pooling, and a fully connected layer (128 units). For CIFAR-10, we use three convolutional layers (32, 64, 128 filters), max pooling, and a fully connected layer (128 units). More information about model architecture is available in appendix \ref{app:model_arch}
SGD is used as the optimizer. For the MNIST dataset, a learning rate of 0.001 with momentum of 0.5 is employed. For Fashion-MNIST, we use a learning rate of 0.001 with momentum of 0.9, and for CIFAR10, the learning rate is set to 0.01 with momentum of 0.9. To ensure model heterogeneity, all local models are initialized with different random states to ensure model initialization heterogeneity. During each communication round, we allow 5 epochs of local training with a batch size of 100. $\beta$ is set to $1$ in Equation \ref{eq:hessian_update} to ensure that the most recent Hessian updates are fully incorporated.

\subsubsection{Baselines}
The proposed method is compared with the following standard baselines:

\textbf{DecHetero}: A decentralized adaptation of FedAvg~\cite{McMahan2016CommunicationEfficientLO}, incorporating the weighted aggregation mechanism described in Equation \ref{eq:fedavg_aggr}. The term \textit{Hetero} highlights that devices start with different initializations, in contrast to the shared initialization assumed by FedAvg.
\textbf{CFA} \cite{Savazzi2019FederatedLW}: CFA, proposed for DFL, takes a distinct approach from DecHetero. Unlike DecHetero, CFA updates its model independently based on individual differences between its neighbors, in order to achieve model consensus across devices.
\textbf{DecDiff} and \textbf{DecDiff (VT)} \cite{Valerio2023CoordinationfreeDF}: DecDiff is another averaging-based approach for DFL  that specifically addresses the model's parameters and data heterogeneity. It contrasts both sources of heterogeneity by incorporating an aggregation function that updates the local model proportionally to its Euclidean distance from the neighborhood-wise aggregated model. The \emph{VT} (i.e., Virtual Teacher) variation of DecDiff uses soft labels computed on the local dataset as a sort of virtual teacher based on the KL loss between the local logits and the soft labels. \textbf{DESA} \cite{huang2024overcoming}: We include DESA, a KD-based approach designed for DFL. DESA introduces synthetic data (``anchors'') generated using distribution matching techniques. These anchors are shared among clients to facilitate mutual learning. As discussed in Section \ref{ssec:desa}, a key aspect of DESA is that the synthetic datasets generated from local data are shared globally and are accessible to all devices in the network. This assumption is reasonable within the peer-to-peer communication framework considered in the original paper. However, DESA’s system assumptions differ significantly from those in this work, where we assume a static graph topology with communication restricted to 1-hop neighborhoods. Despite these differences, we include a comparison in a dedicated section to provide a comprehensive analysis and to demonstrate how constrained, real-world topologies significantly impact DFL performance.

\begin{figure*}[t]
    \centering
    \subfloat[]{\includegraphics[width=0.3\textwidth]{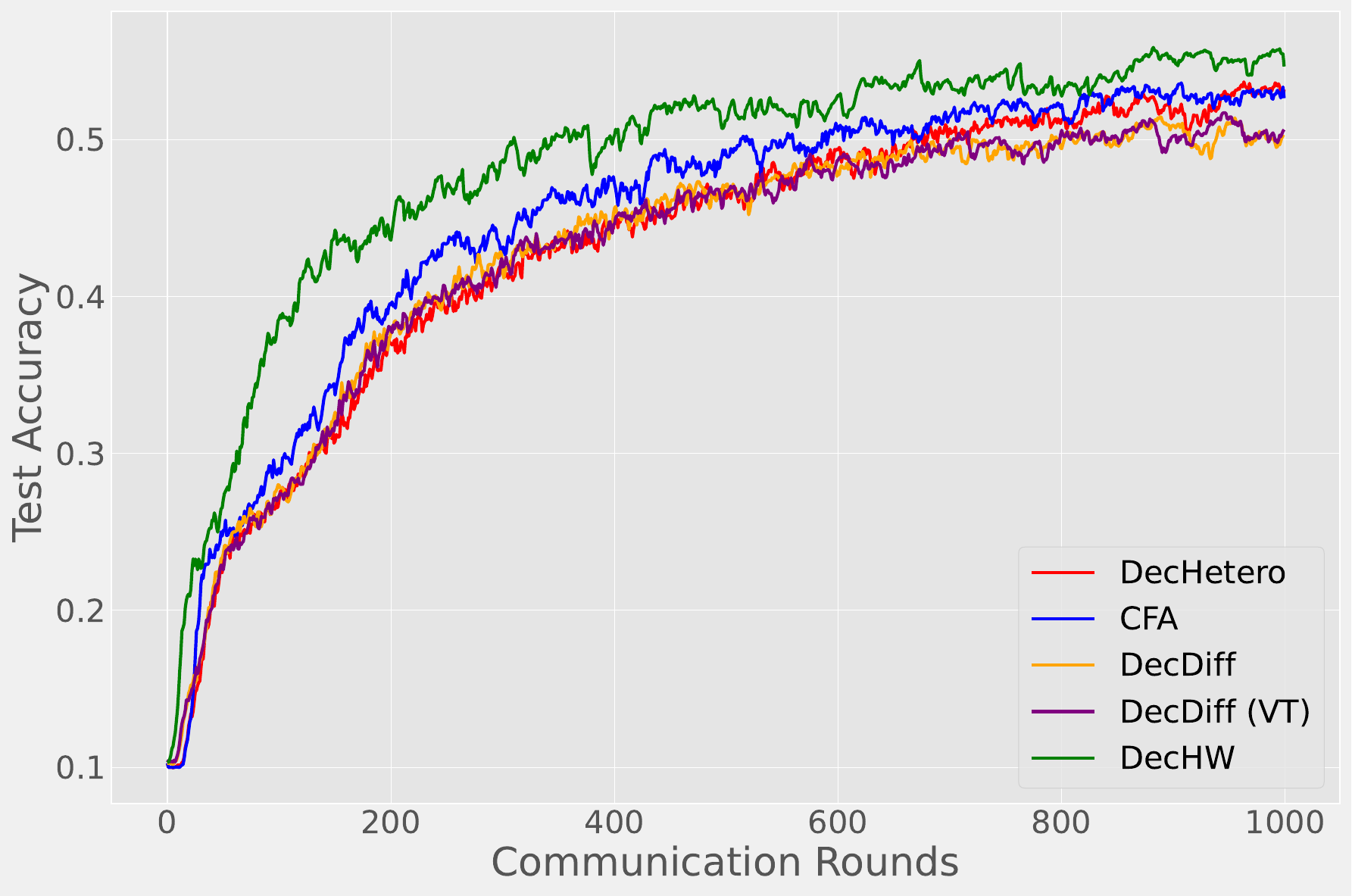}}
    \subfloat[]{\includegraphics[width=0.3\textwidth]{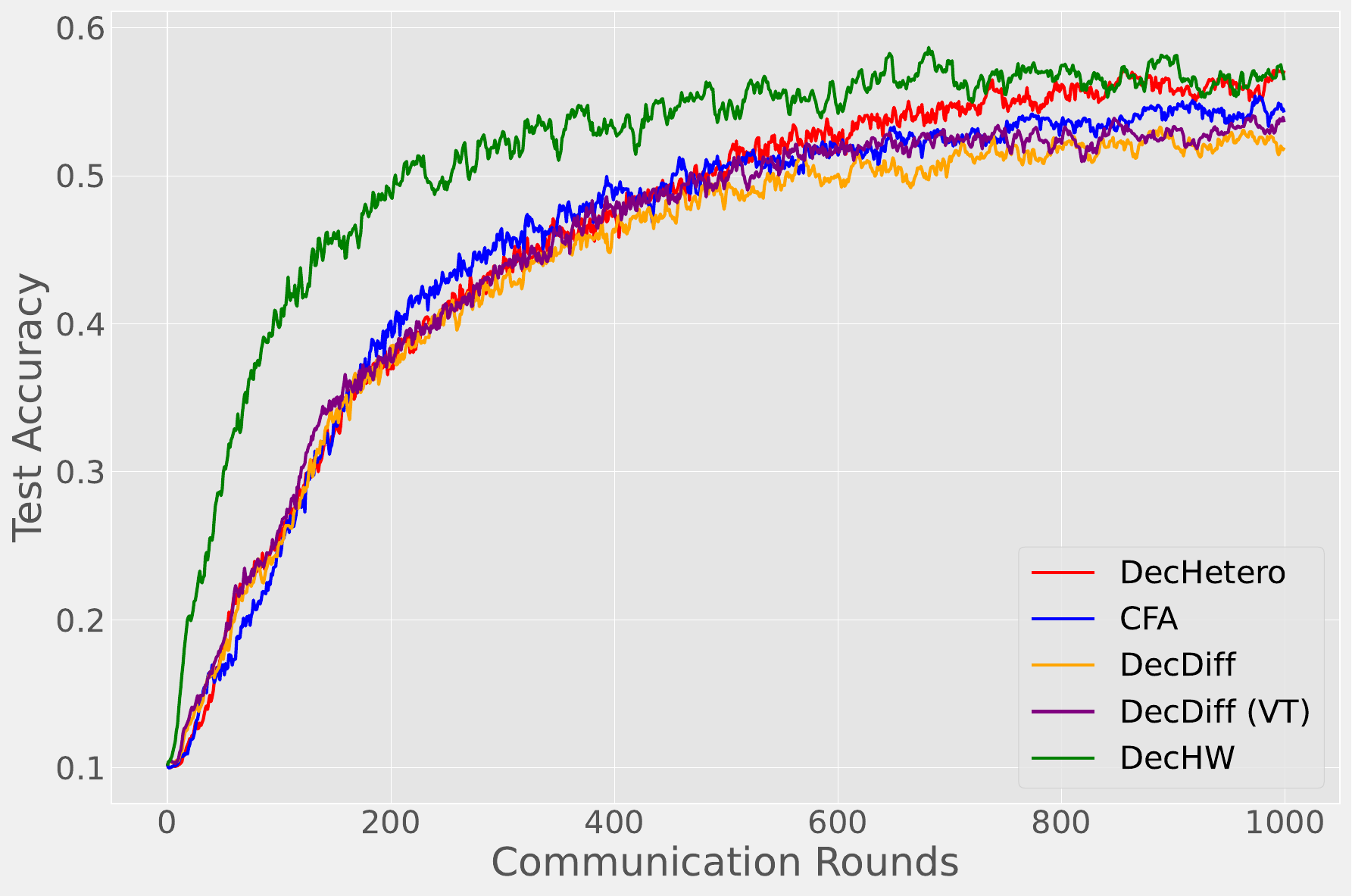}}
    \subfloat[]{\includegraphics[width=0.3\textwidth]{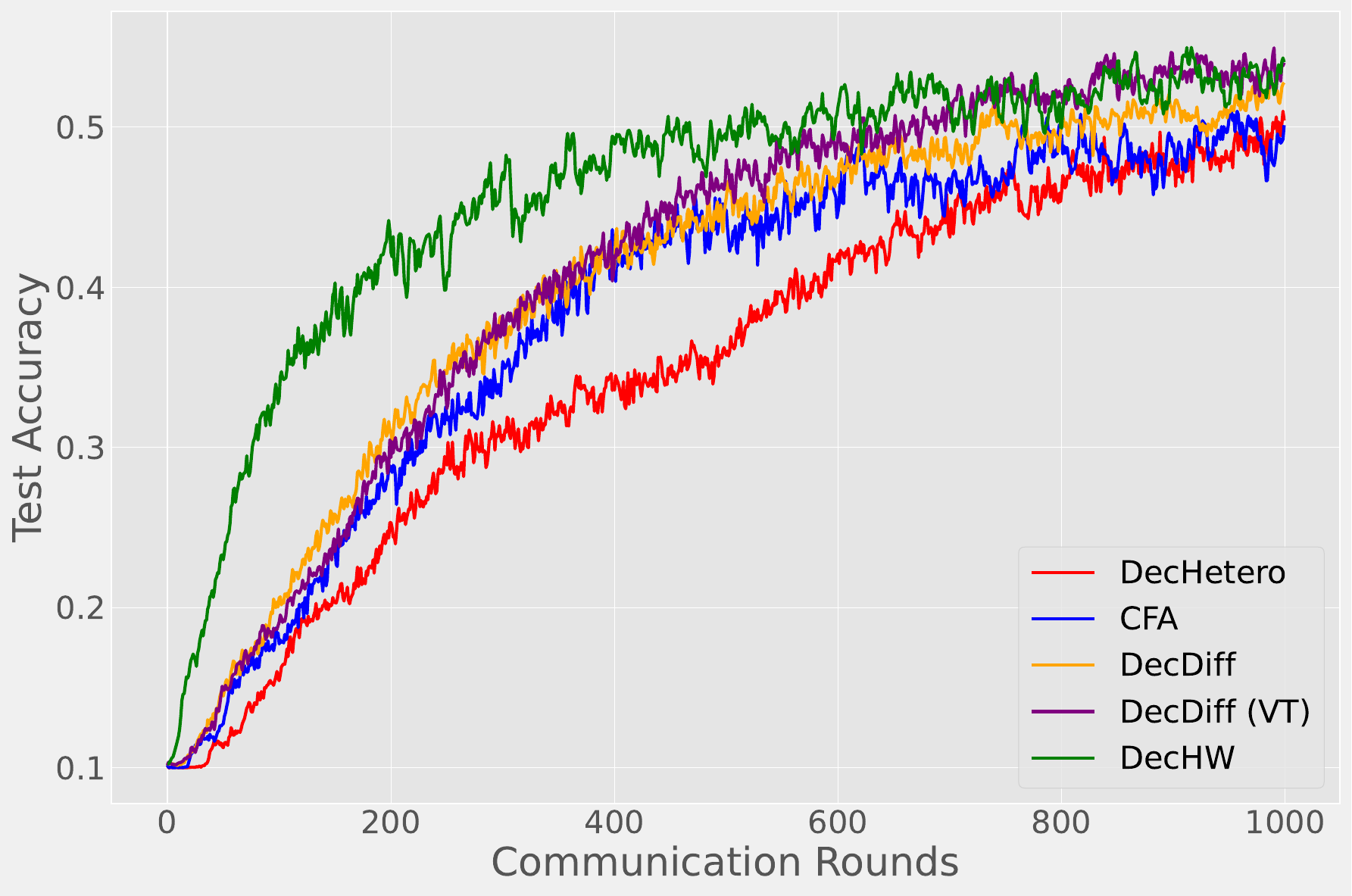}}\\
    \subfloat[Dir(1)]{\includegraphics[width=0.3\textwidth]{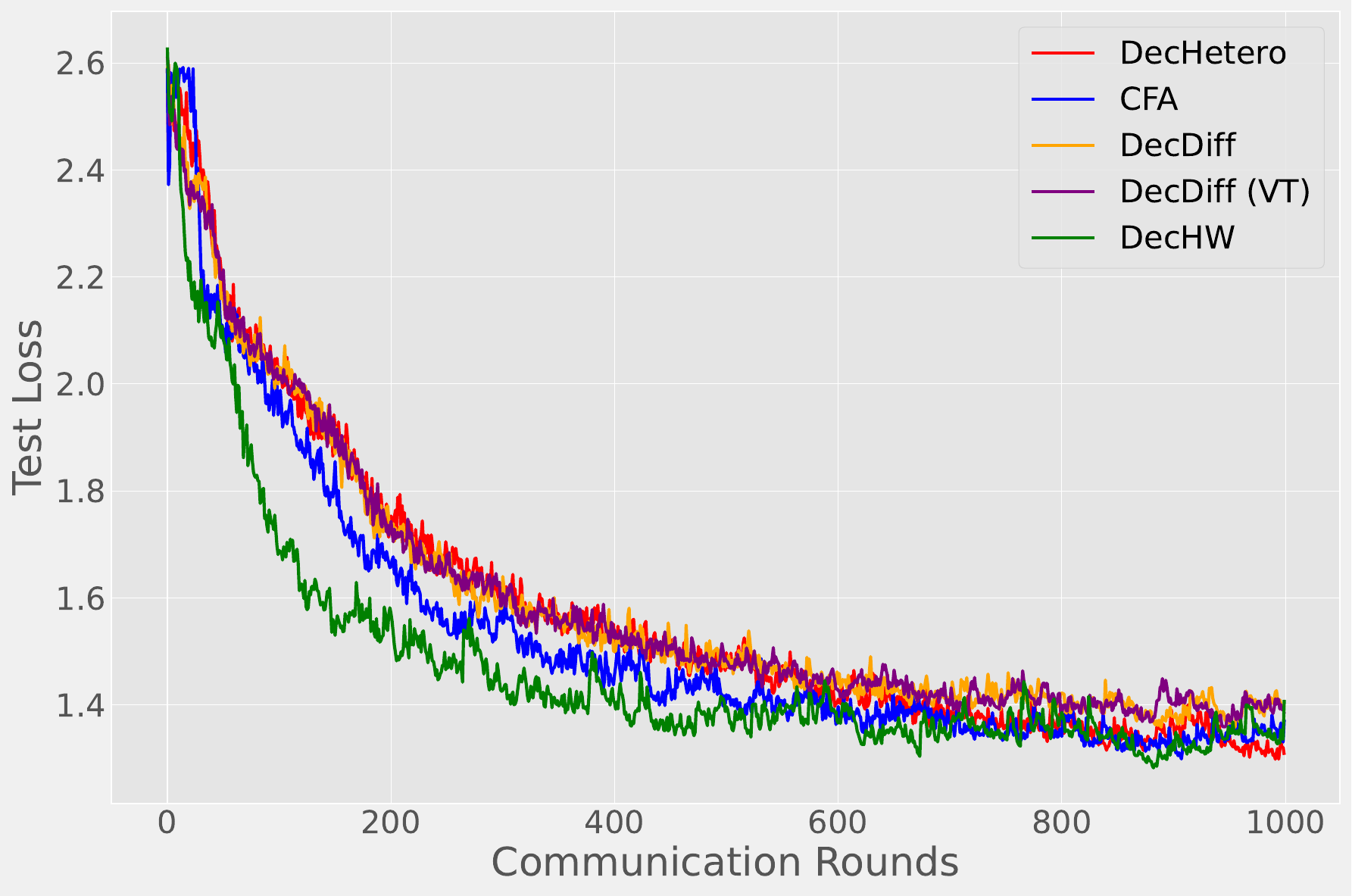}\label{fig:data_dir1}}
    \subfloat[Dir(0.5)]{\includegraphics[width=0.3\textwidth]{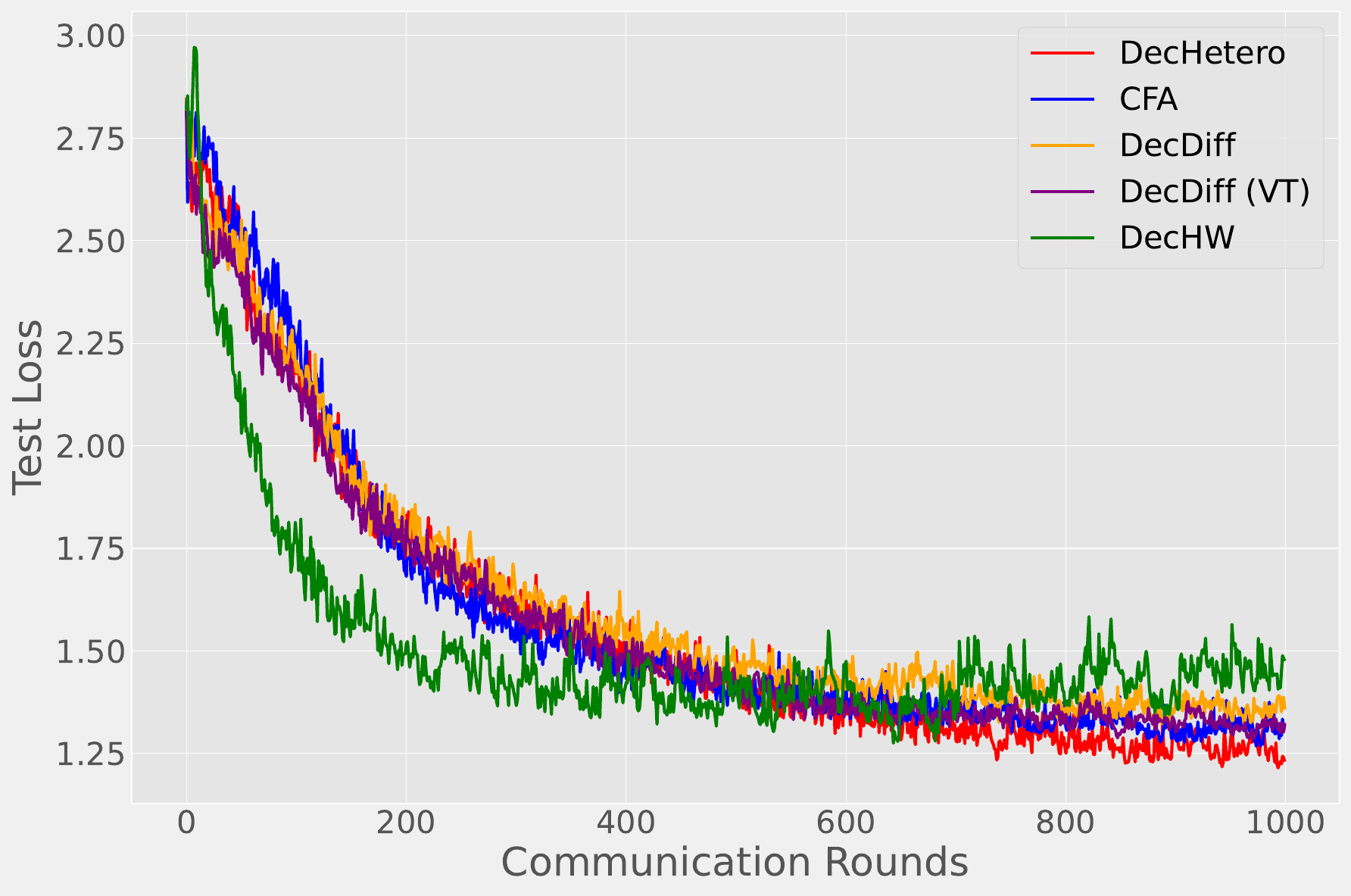}\label{fig:data_dir0.5}}
    \subfloat[Dir(0.2)]{\includegraphics[width=0.3\textwidth]{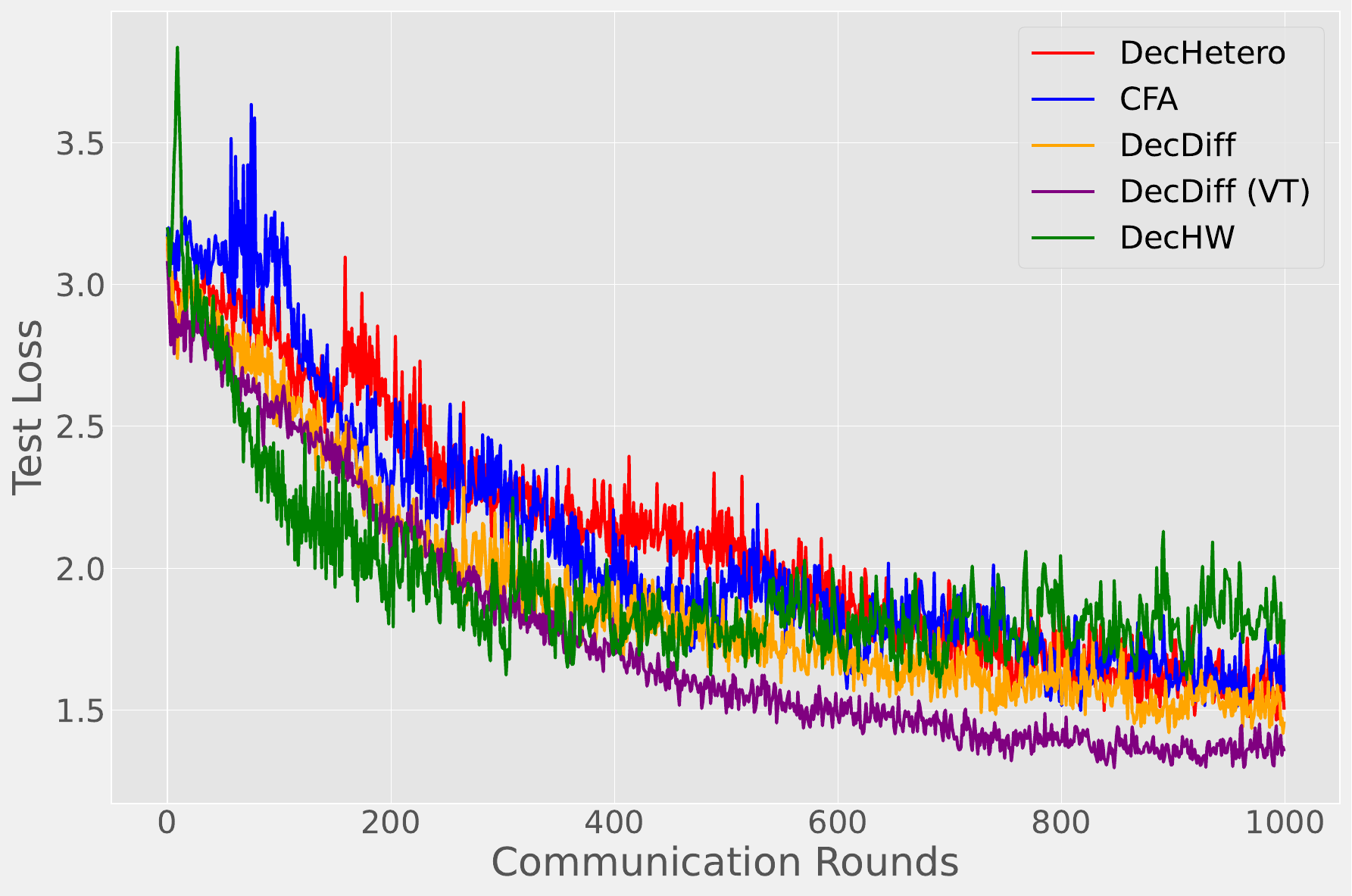}\label{fig:data_dir0.2}}

    \caption{ Average accuracy and loss across devices on CIFAR10 dataset. Mean values from 5 different runs are reported as a function of communication rounds.}
    \label{fig:cifar10_DFL}
\end{figure*}



\begin{table}[]
    \centering
    \caption{Mean $\pm$ Std of test Accuracy in last communication round when CNN is trained on CIFAR10 dataset.}
    \resizebox{0.48\textwidth}{!}{%
    \begin{tabular}{cccc}
         \toprule
         Method & Dir(1) & Dir(0.5) & Dir(0.2)\\
         \midrule
         DESA & $35.18 \pm 0.008$ & $30.10 \pm 0.010$ & $23.46 \pm 0.002$\\
         DecHetero & $52.81 \pm 0.008$ & $56.67 \pm 0.023$ & $51.28 \pm 0.027$ \\
         DecHW & $\bm{55.29 \pm 0.019}$ & $\bm{57.16 \pm 0.018}$ & $\bm{54.37 \pm 0.020}$\\
         \bottomrule
    \end{tabular}
    }
    \label{tab:desa}
\end{table}

\subsection{Results} \label{sec:results}
We conducted experiments for 1000 communication rounds. After each round, we computed the mean accuracy across all devices on the test data. The results are reported as Mean $\pm$ Std over $5$ different runs.

Table \ref{tab:accuracy} presents the average accuracy achieved in the last communication round with the best performer highlighted in bold. DecHetero performs reasonably well across all datasets when $\alpha = 1$. Notably, $\alpha = 1$ simulates less heterogeneous data, resulting in faster model convergence. However, as the degree of data heterogeneity increases, DecHetero's performance degrades by $1.53 \text{ to } 5.08$ percentage points across different datasets. Similarly, CFA and DecDiff tend to struggle under high levels of data heterogeneity. DecDiff (VT) outperforms the other baselines on MNIST and Fashion-MNIST, but its performance degrades as the complexity of the CIFAR-10 dataset increases.
On the other hand, the proposed approach, DecHW, consistently outperforms all other methods across all datasets and under various levels of data heterogeneity. This demonstrates its ability to scale effectively across different types of datasets.

\begin{figure}
    \centering
    \includegraphics[width=0.85\linewidth]{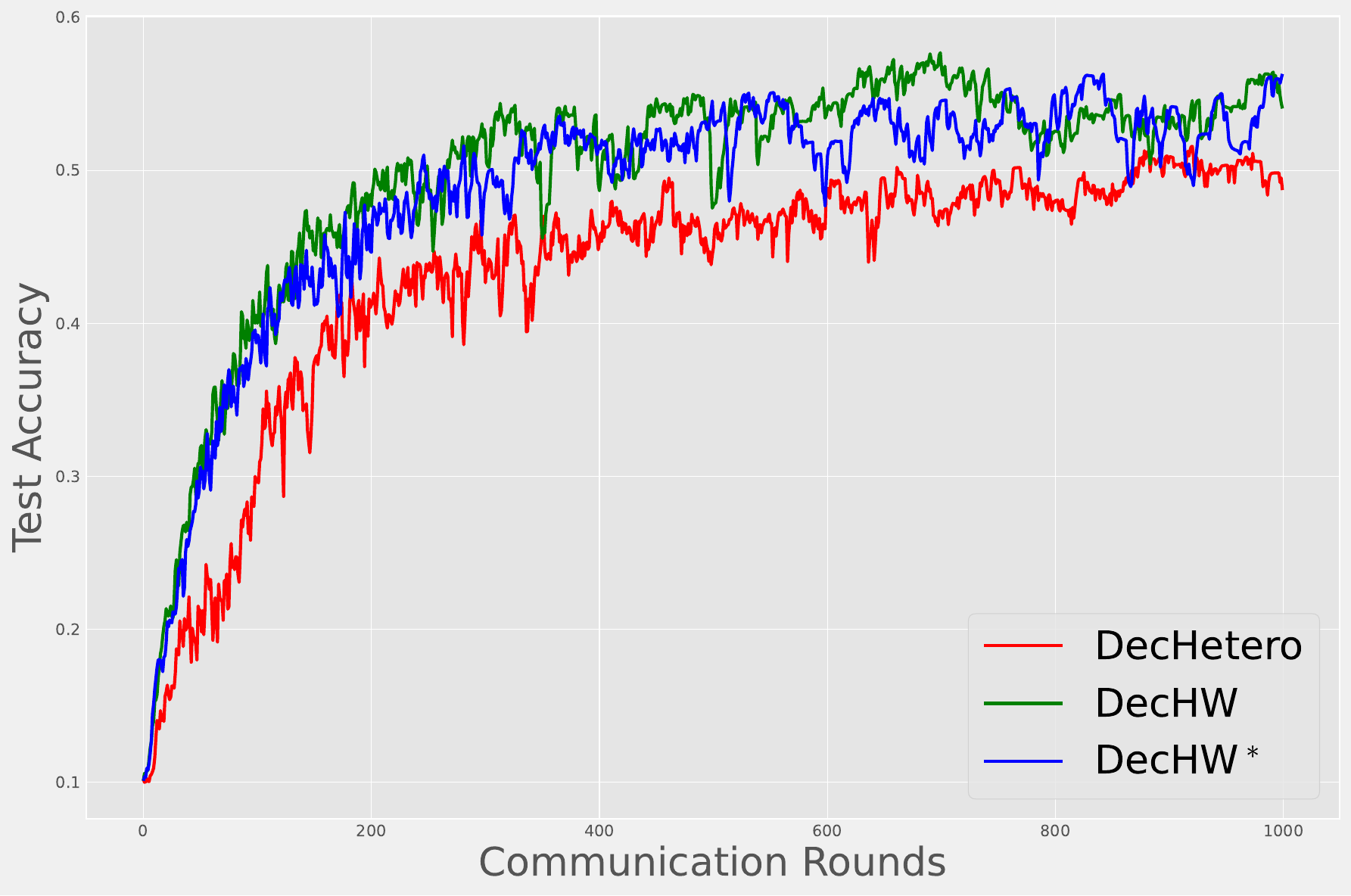}
    \caption{DecHW is compared with its variant DecHW$^*$ on the CIFAR10 dataset.} \vspace{-10pt}
    \label{fig:efficient_dechw}
\end{figure}

We now focus on the robustness of the proposed method in terms of convergence time. Table \ref{tab:communication_effic} presents the total number of communication rounds each method requires to reach a specific level of accuracy, as reported in Table \ref{tab:accuracy}. The best performer is highlighted in bold. Due to the heterogeneous initialization of local models, different methods struggle in the early communication rounds to converge toward the generalized model. In these early rounds, there is a significant risk of information loss when aggregating models using naive strategies, as there is no one-to-one neuron matching, and averaging local updates results in substantial information degradation. Consequently, it takes several rounds for information to diffuse throughout the network topology. In Table \ref{tab:communication_effic}, methods such as DecHetero and CFA naively aggregate local updates and require several rounds to achieve a certain level of accuracy. DecDiff attempts to address this by considering model-level importance through a regularization term, which leads to faster convergence. DecHW, on the other hand, explicitly addresses this issue by considering the parameter-level importance of each local model. In the case of extreme data heterogeneity for CIFAR10, while other approaches require more than $150$ rounds to reach $50\%$ of the maximum accuracy, DecHW manages to reach this milestone in only $59$ rounds. This significant reduction in communication rounds saves substantial amounts of time and mutual communication between devices. As a result, the proposed method is not only robust in achieving high accuracy but also demonstrates fast convergence. Moreover, Figure~\ref{fig:cifar10_DFL} shows the learning curves on the CIFAR10 dataset. Proposed approach DecHW consistently achieves fast convergence from early rounds of communication and also remains top performer throughout the training process. Variants of Table \ref{tab:communication_effic} with different reference point accuracies and the learning curves for the other datasets can be found in the appendix \ref{app:further_exp}

\subsubsection{DecHW vs DESA}  
\label{ssec:desa}
In Table~\ref{tab:desa}, we present the mean accuracy achieved by DecHW and DESA~\cite{huang2024overcoming} on CIFAR10, the most challenging of the considered datasets. Unlike DecHW and the policies discussed earlier, which rely solely on the original local data without any sharing, DESA introduces a key innovation: the generation of a shared synthetic dataset to address the heterogeneity inherent in DFL. While DESA has proven highly effective in scenarios assuming a peer-to-peer network with a complete communication graph (i.e., any device can communicate with any other device) and where communication occurs with a randomly selected subset of devices at each round, real-world DFL typically involves significant constraints on communication. These constraints may arise from trust issues or limited communication opportunities, resulting in much smaller and less connected neighborhoods.

To evaluate DESA under these more realistic conditions, we implemented it using our topology, where node neighborhoods are considerably smaller than the full graph. As shown in Table~\ref{tab:desa}, the effectiveness of DESA’s synthetic dataset diminishes significantly when nodes cannot communicate freely with any other node over time. Under this constrained topology, DESA incurs a substantial performance penalty and is outperformed even by the simplest averaging strategy.

\subsubsection{Communication cost}

In Tables~\ref{tab:accuracy} and~\ref{tab:communication_effic}, we imposed no restrictions on the use of second-order information to demonstrate the unconstrained performance of DecHW. However, leveraging second-order information requires additional communication between neighbors, beyond the standard exchange of model parameters. Specifically, it necessitates sharing the diagonal of the Hessian, which has the same size as the model itself. As shown in Figure~\ref{fig:hess_norm}, the utility of the Hessian information diminishes rapidly, dropping to near zero after only a few (tens of) communication rounds.

In this subsection, we demonstrate that using Hessian information for a limited number of communication rounds before transitioning to standard averaging achieves comparable results to using the Hessian throughout the entire training process. To evaluate this, we set up a particularly challenging scenario using the CIFAR10 dataset and limiting the use (and exchange) of the Hessian information to only the first 30 communication rounds. Notably, as shown in Table~\ref{tab:communication_effic}, DecHW has yet to achieve 50\% accuracy for any data distribution by communication round 30.

In Figure~\ref{fig:efficient_dechw}, we compare the standard DecHW against a modified version (denoted as DecHW$^*$), where the exchange of Hessian information ceases after 30 communication rounds. The results indicate that the performance—both over time and at steady-state—is nearly identical between the two approaches. This underscores that incorporating Hessian information during the initial phase of the learning process (in this case, 30 communication rounds out of 1,000) is sufficient to capture its benefits.

Finally, we compare the communication overhead incurred during $T$ communication rounds (with $T=1,000$ in our simulations) when the Hessian is used for $\theta$ rounds (where $\theta=30$ in our experiment). For DecHetero, the information exchange is limited to $T \times \text{modelSize}$ parameters per round. In contrast, DecHW requires $T \times 2 \times \text{modelSize}$, while DecHW$^*$ involves $T \times \text{modelSize} + \theta \times \text{modelSize}$. For small values of $\theta$, the additional overhead from exchanging Hessian information is minimal, making this approach highly efficient.

\vspace{-10pt}
\section{Conclusion}

This paper introduces DecHW, an effective aggregation method for DFL designed to address challenges posed by data heterogeneity and variations in local model initializations. DecHW leverages second-order information from local models to capture parameter-level variations, enhancing the aggregation process. Specifically, the diagonal of the Hessian matrix is utilized to quantify the sensitivity of each parameter to the local data distribution, enabling the scaling of parameters based on their local importance during aggregation. Experimental results demonstrate that DecHW achieves fast convergence across different datasets, even under varying degrees of data heterogeneity.


\appendices
\section{Data allocations}\label{app:data_dist}
\begin{figure}[t]
    \centering
    \subfloat[Dir(1)]{
        \includegraphics[width=0.95\columnwidth]{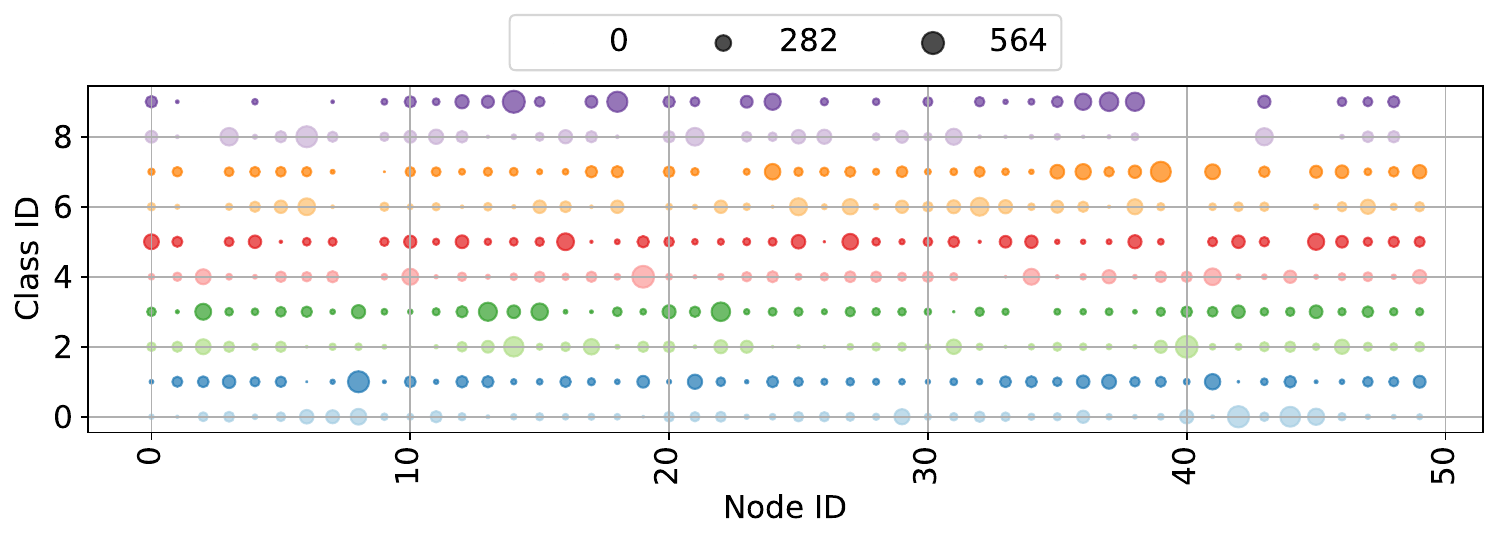}
        \label{fig:dir1}
    }

    \vspace{0.5em} 

    \subfloat[Dir(0.5)]{
        \includegraphics[width=0.95\columnwidth]{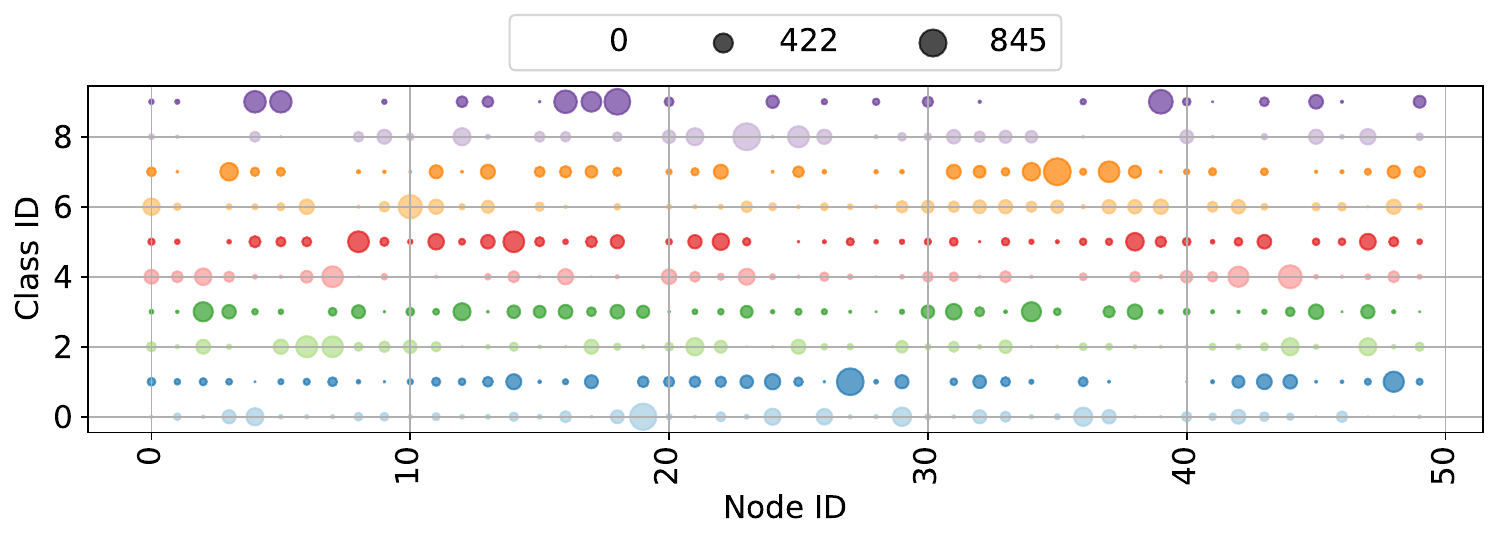}
        \label{fig:dir0.5}
    }

    \vspace{0.5em}

    \subfloat[Dir(0.2)]{
        \includegraphics[width=0.95\columnwidth]{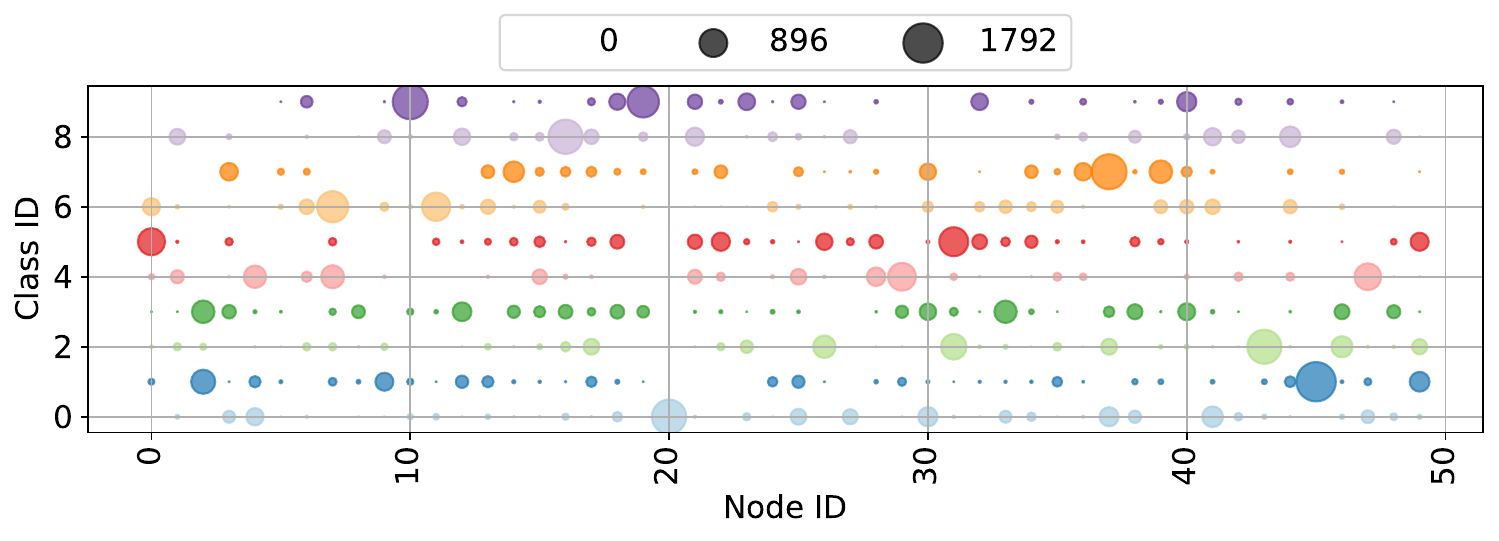}
        \label{fig:dir0.2}
    }
    \caption{Data allocation across devices using Dirichlet distributions. The size of each circle indicates the quantity of data allocated from each class to a node.}
    \label{fig:dir_dist}
\end{figure}
We use a Dirichlet distribution with $\alpha = \{0.2, 0.5, 1\}$ to simulate the non-IID data distribution across the devices. Lower values of $\alpha$ result in greater data heterogeneity, whereas higher values of $\alpha$ reduce heterogeneity, producing a more balanced distribution. For every class label, a random draw from the Dirichlet distribution is performed, and the corresponding multinomial distribution dictates how many training examples are allocated to each client. As a result, the value of $\alpha$ directly governs the degree of heterogeneity: smaller values create highly heterogeneous scenarios, while larger values yield more uniform data distributions. The data allocation across devices is shown in Figure~\ref{fig:dir_dist}.

\section{Model architecture and configuration} \label{app:model_arch}
Table \ref{tab:nn_models} presents the model architectures used in the experiments. For MNIST, the CNN consists of two convolutional layers with 10 and 20 filters, respectively, each using a \(5 \times 5\) kernel, followed by a max-pooling layer and two dense layers containing 320 and 50 units. In the case of Fashion-MNIST, the network comprises two convolutional layers with 32 and 64 filters of size \(3 \times 3\), a max-pooling layer, and a dense layer with 128 units. For CIFAR-10, the architecture includes three convolutional layers with 32, 64, and 128 filters, respectively, followed by a max-pooling layer and a dense layer with 128 units.

\begin{algorithm}[h]
    \caption{DecHW Algorithm}
    \label{alg:dechw}
    \begin{algorithmic}[1] 
    \REQUIRE {$\mathcal{N}_i$: set of devices in the neighborhood of device $i$; $D_i$: local dataset for devices$i$;  $w_i$ model's parameters for device $i$; $t$: is the $t$-th communication round; $E$: local training epochs; $T$: total communication rounds; $n$ is the $n$-th parameter of $w_i$; $m$ is the size of $w_i$}
    \STATE Initialize local model $w_i$ at random for each device $i$
    \STATE Each device $i$ computes $\mathrm{train}(w_i^{(0)},D_i)$
    \STATE Each device $i$ computes $\hat{H}_{i,\text{diag}}^{(0)} = \frac{H_{i,\text{diag}}}{\|H_{i,\text{diag}}\|_2}$
    \STATE Each device communicates $\{w_i^{(0)},\hat{H}_{i,\text{diag}}^{(0)}\}$ to its neighbors in $\mathcal{N}_i$ 
    \FOR{$t=1, 2, \dots, T$}
        \STATE {\textbf{on each device} $i\in\mathcal{N}$ \textbf{in parallel do}}
        \STATE {compute $p_{i,n}^{(t)} = \frac{\hat{H}_{i,n}^{(t-1)}}{\sum_{j=1}^{N} \hat{H}_{j,n}^{(t-1)}}$}
    \STATE {$\forall n=1,\dots,m$ perform aggregation $$w_{i,n}^{(t)} =
\begin{cases}
\sum_{j=1}^{N} p_{j,n}^{(t)} w_{j,n}^{(t-1)} & \text{if } \sum_{j=1}^{N} p_{j,n}^{(t)} \neq 0 \\
 \sum_{j=1}^{N} \tau_j w_j^{(t-1)} & \text{otherwise}
\end{cases}$$}
    \STATE $w_i^{(t)} = \mathrm{train}(w_i^{(t)},D_i)$
    \STATE {compute $\hat{H}_{i,\text{diag}}^{(t)} = \hat{H}_{i,\text{diag}}^{(t-1)} +  \beta \frac{H_{i,\text{diag}}^{(t)}}{\|H_{i,\text{diag}}^{(t)}\|_2}$}
    \STATE {communicate $\{w_i^{(t},\hat{H}_{i,\text{diag}}^{(t)}\}$ to all $j \in \mathcal{N}_i$}
\ENDFOR
\end{algorithmic}
\label{algo:pseudo}
\end{algorithm}

\section{DecHW Pseudocode}\label{app:pseudo}

\begin{figure*}[t]
  \centering
  \subfloat[Accuracy]{%
    \includegraphics[width=0.45\textwidth]{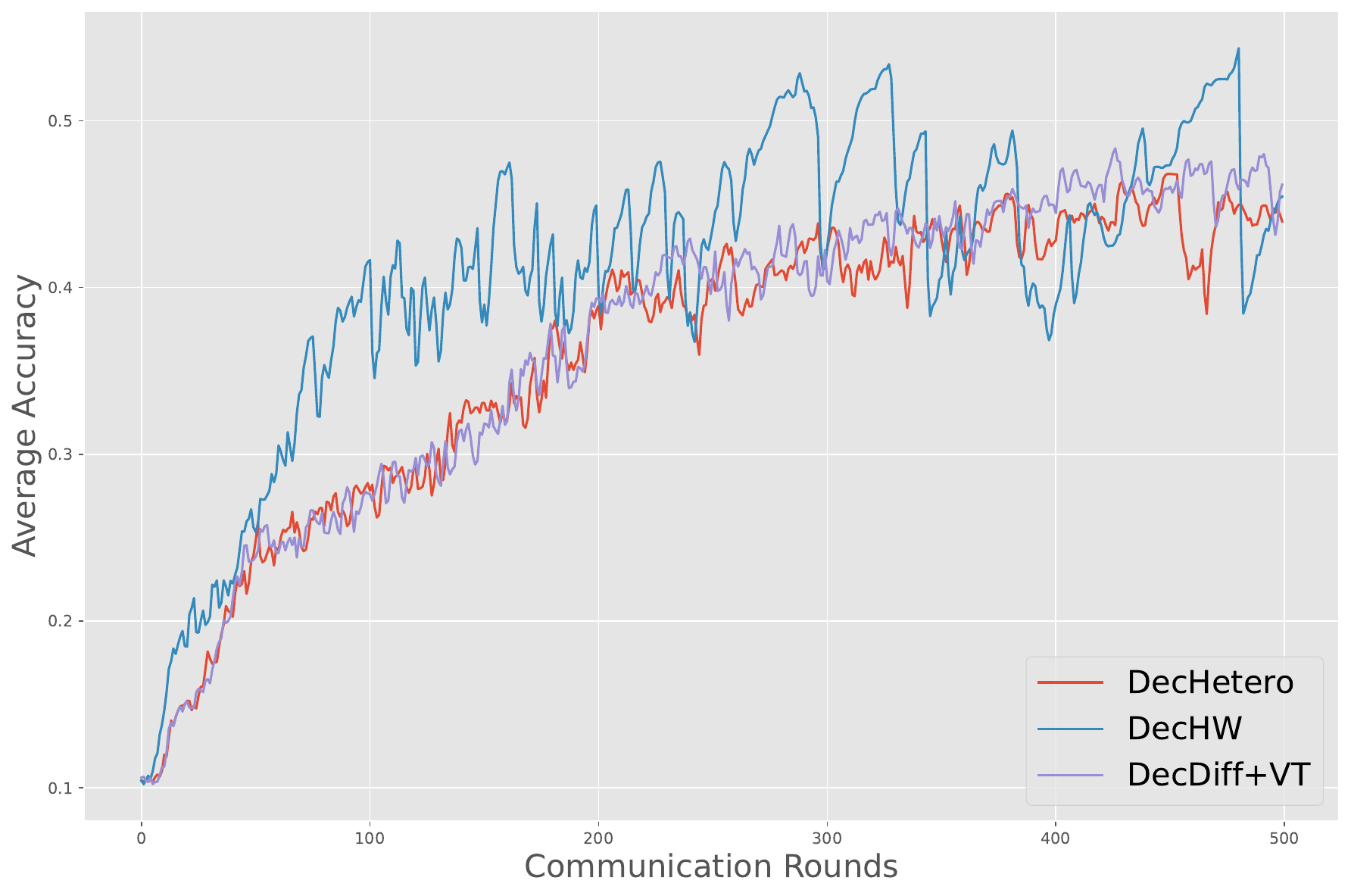}
    \label{fig:non_accum_acc}
  }
  \hfill
  \subfloat[Loss]{%
    \includegraphics[width=0.45\textwidth]{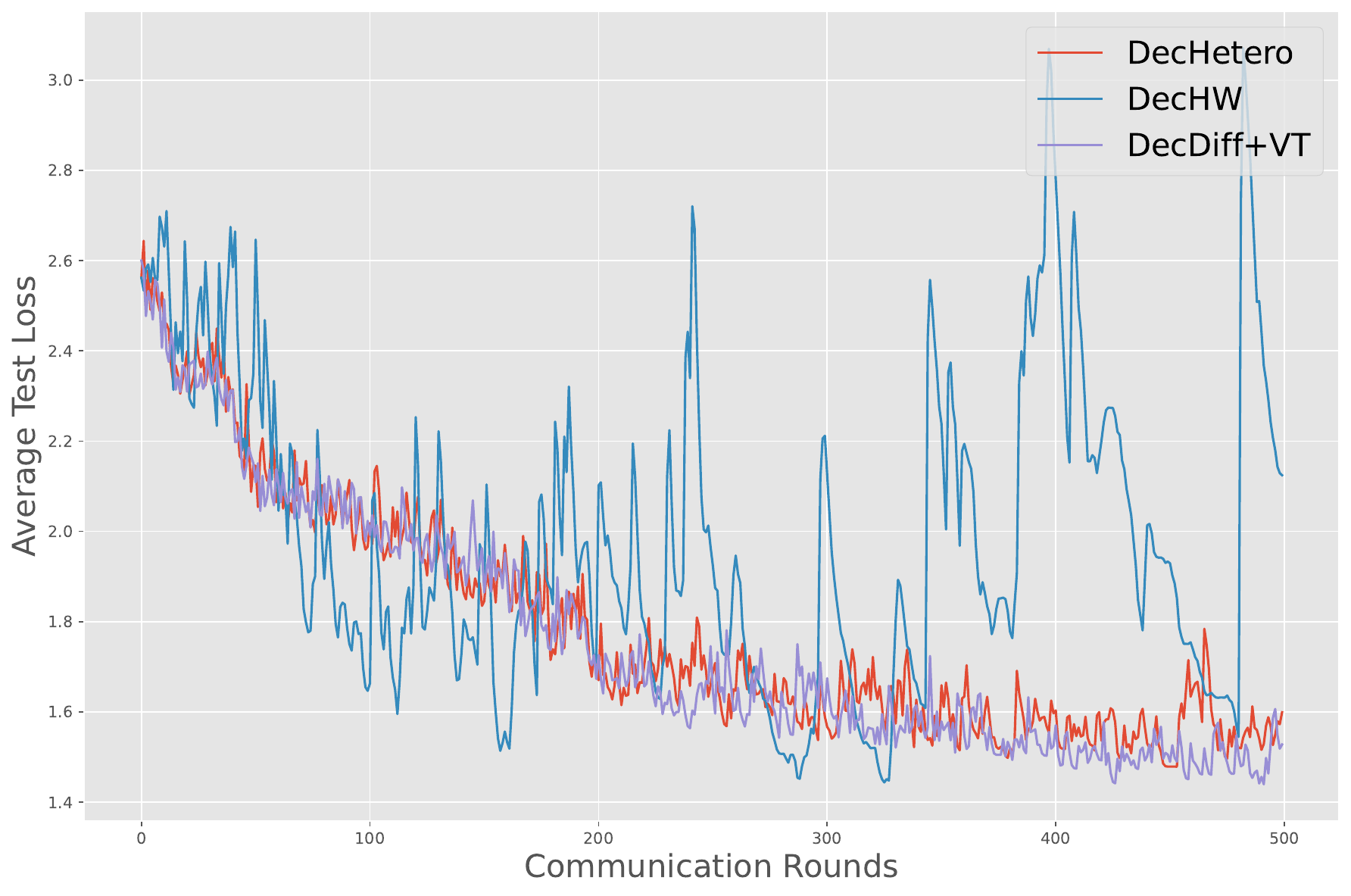}
    \label{fig:non_accum_loss}
  }
  \caption{Performance of DecHW on CIFAR-10 dataset without accumulating past Hessian diagonals: Only the current round's Hessian diagonal is used to compute aggregation weights, without incorporating information from previous communication rounds.}
  \label{fig:non_accum_hess}
\end{figure*}

\begin{figure*}[t]
  \centering
  \subfloat[Hessian diagonal accumulation]{%
    \includegraphics[width=0.45\textwidth]{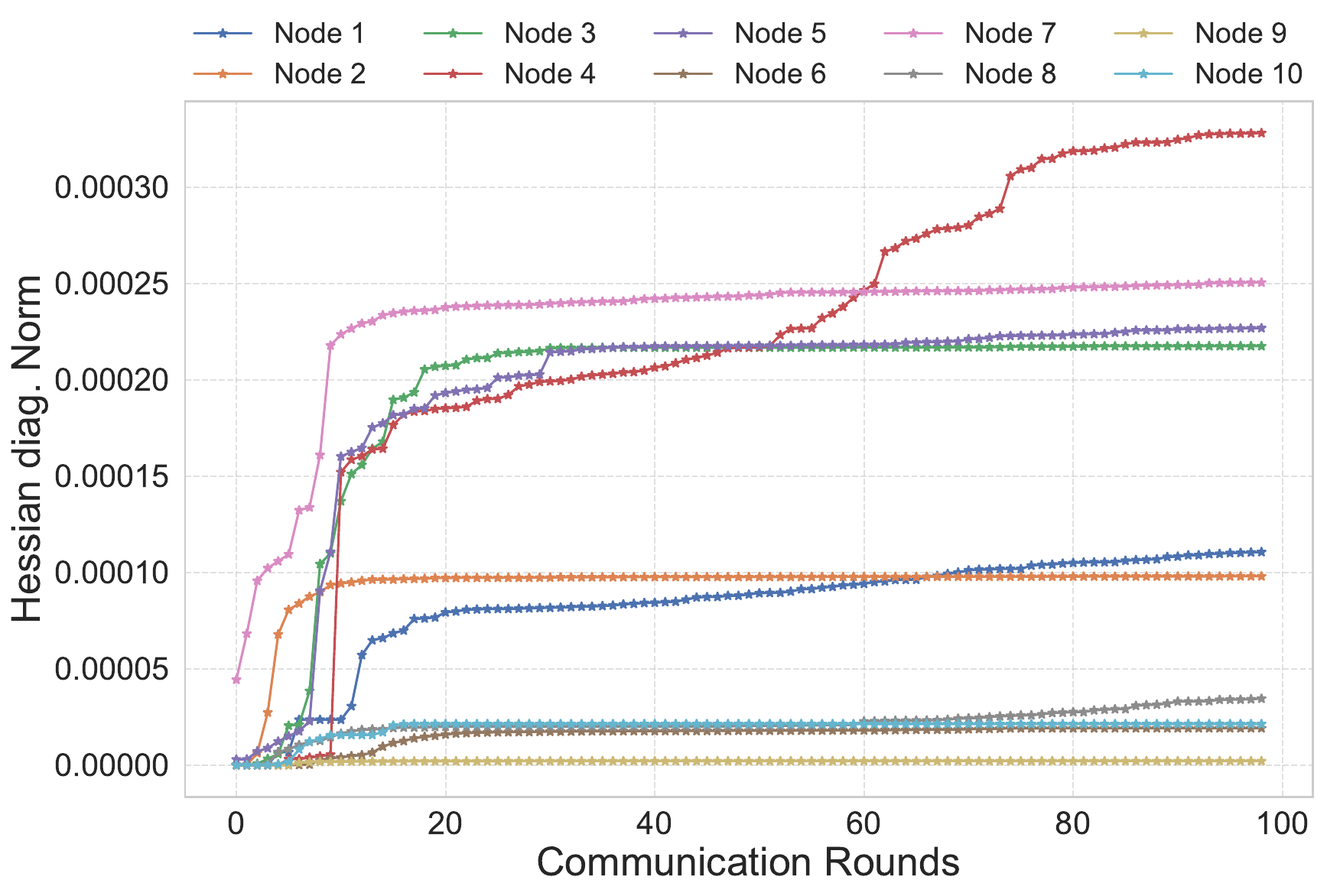}
    \label{fig:hessian_accum}
  }
  \hfill
  \subfloat[Resultant weights]{%
    \includegraphics[width=0.45\textwidth]{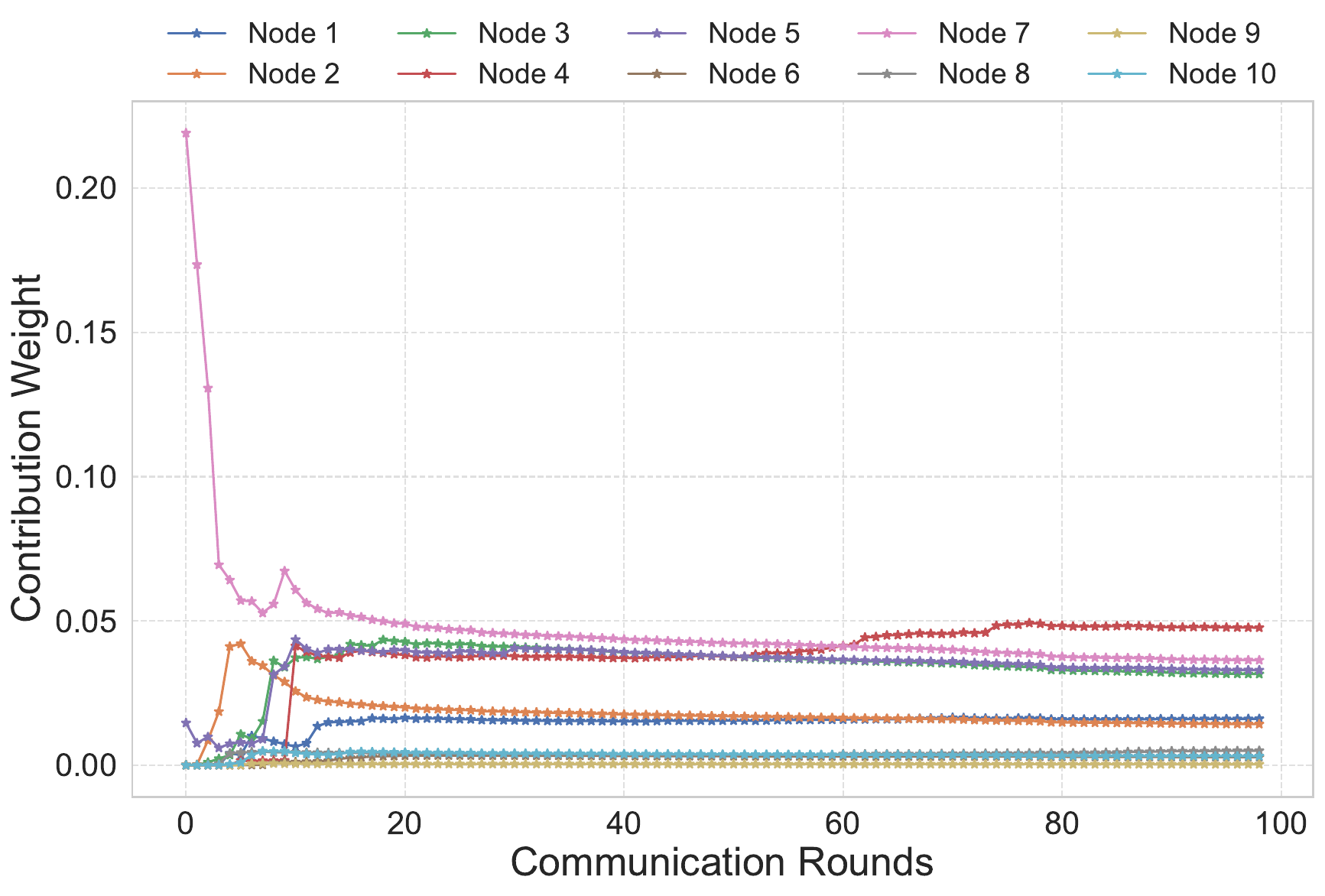}
    \label{fig:weight_consistency}
  }
  \caption{Stability of Hessian-weighted aggregation over communication rounds. (a) shows the accumulation of a selected Hessian diagonal element across 10 clients, where values initially increase but later stabilize—indicating convergence to flat regions of the loss surface. (b) shows the corresponding aggregation weights, which remain stable across rounds, preserving the influence of sensitive parameters without introducing instability.}
  \label{fig:hessian_stability}
\end{figure*}
Algorithm~\ref{alg:dechw} presents the procedural steps for our proposed DecHW method. In each communication round, participating devices exchange both their local model parameters and the normalized diagonal of the Hessian matrix with their neighbors. Aggregation is performed in a parameter-wise fashion, where weights are derived from the relative sensitivity of each parameter, as indicated by the normalized Hessian diagonals. 

The algorithm includes local model training, per-parameter weighted aggregation, and a moving average update of the Hessian diagonal to smooth temporal fluctuations. This design enables DecHW to adapt to heterogeneous data distributions while maintaining stable and efficient learning in fully decentralized environments.
\begin{table}[t]
    \centering
    \caption{Models Architectures}
    \label{tab:nn_models}
    \resizebox{0.5\textwidth}{!}{%
    \begin{tabular}{ccccc}
        \toprule
        Dataset & Model & Layers & Activation & Kernel Size  \\
        \midrule
        \multirow{3}{*}{MNIST} & \multirow{3}{*}{CNN} & Conv2d: 10, 20& ReLU & $5 \times 5$ \\
         & & MaxPool(2)&   &  \\
        &  & FC: 320,50&  ReLU &  \\               
        \midrule
        \multirow{3}{*}{Fashion} & \multirow{3}{*}{CNN} & Conv2d: 32, 64& ReLU & $3 \times 3$ \\
         & & MaxPool(2)&   &  \\
        &  & FC: 128&  ReLU &  \\               
        \midrule
        \multirow{3}{*}{CIFAR10} & \multirow{3}{*}{CNN} & Conv2d: 32, 64, 128 & ReLU & $3 \times 3$ \\
         & & MaxPool(2)&   &  \\
        &  & FC: 128&  ReLU &  \\               
        \bottomrule
    \end{tabular}
    }

\end{table}

\section{Importance of Hessian Accumulation in Decentralized Learning}
We first investigate the performance drop of DecHW when Hessian accumulation is disabled, as illustrated in Figure~\ref{fig:non_accum_hess}. In this setting, only the current round's Hessian diagonal is used for computing aggregation weights, discarding historical curvature information. While test accuracy and loss on the CIFAR-10 dataset initially improve, performance quickly destabilizes in later rounds. This instability arises because the Hessian diagonal evolves throughout different phases of training.
In the early stages, model parameters are still evolving, and their local curvature can be approximated from recent gradients. However, as training progresses and parameters begin to converge, the corresponding Hessian diagonal values tend to diminish, often approaching zero, particularly for directions along flatter regions of the loss surface. This trend is clearly captured in Figure~\ref{fig:hess_norm}, where the Hessian norm progressively declines over communication rounds. The lack of accumulated curvature leads to underestimation of parameter importance in later stages, which in turn results in noisy or inconsistent aggregation weights.

In contrast, our proposed approach accumulates Hessian diagonal information over time, as formalized in Equation~\ref{eq:hessian_update}. This cumulative strategy smooths short-term fluctuations and stabilizes curvature estimation across rounds. Figure~\ref{fig:hessian_accum} illustrates that while the accumulated Hessian initially grows, it eventually saturates, indicating that most parameters have reached flatter regions of their local loss landscapes. Consequently, the aggregation weights remain consistent and robust, as shown in Figure~\ref{fig:weight_consistency}, enabling the model to preserve the influence of important parameters during early learning while avoiding instability in later stages. This highlights the critical role of historical curvature in achieving stable and effective federated learning.

This accumulation mechanism also helps reduce computational overhead and communication costs. To explore this, we experiment with early termination of Hessian updates, as detailed in Section~\ref{sec:results} and illustrated in Figure~\ref{fig:efficient_dechw}. The results demonstrate that variants of our method with early termination still achieve comparable performance, validating the robustness of DecHW even when Hessian information is not computed at every communication round. This further encourages efficient and scalable decentralized training without sacrificing stability or accuracy.


\begin{figure*}[b]
    \centering

    \subfloat[Accuracy (Dir(1))]{%
        \includegraphics[width=0.32\textwidth]{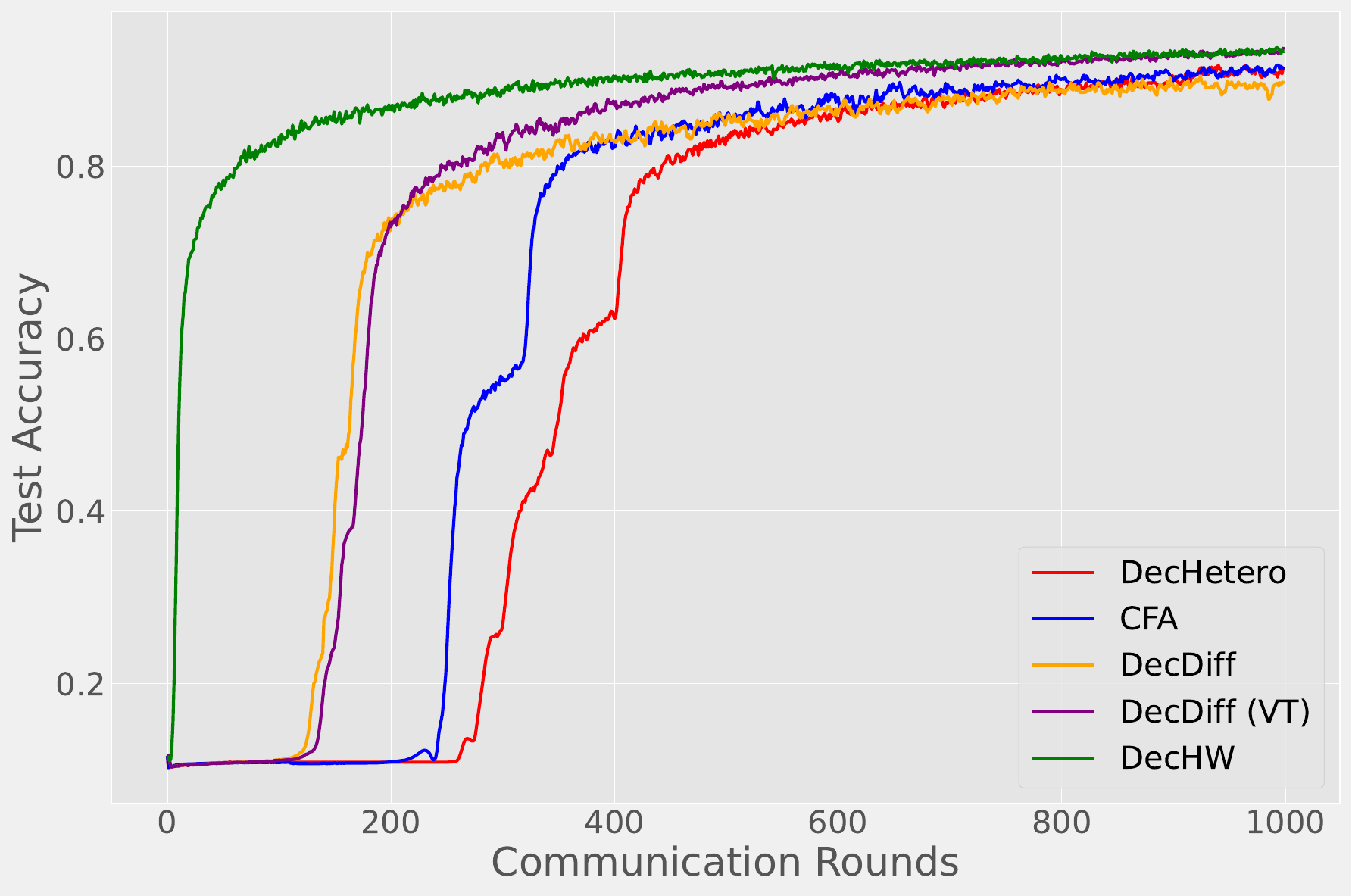}
        \label{fig:mnist_dir1_acc}
    }
    \hfill
    \subfloat[Accuracy (Dir(0.5))]{%
        \includegraphics[width=0.32\textwidth]{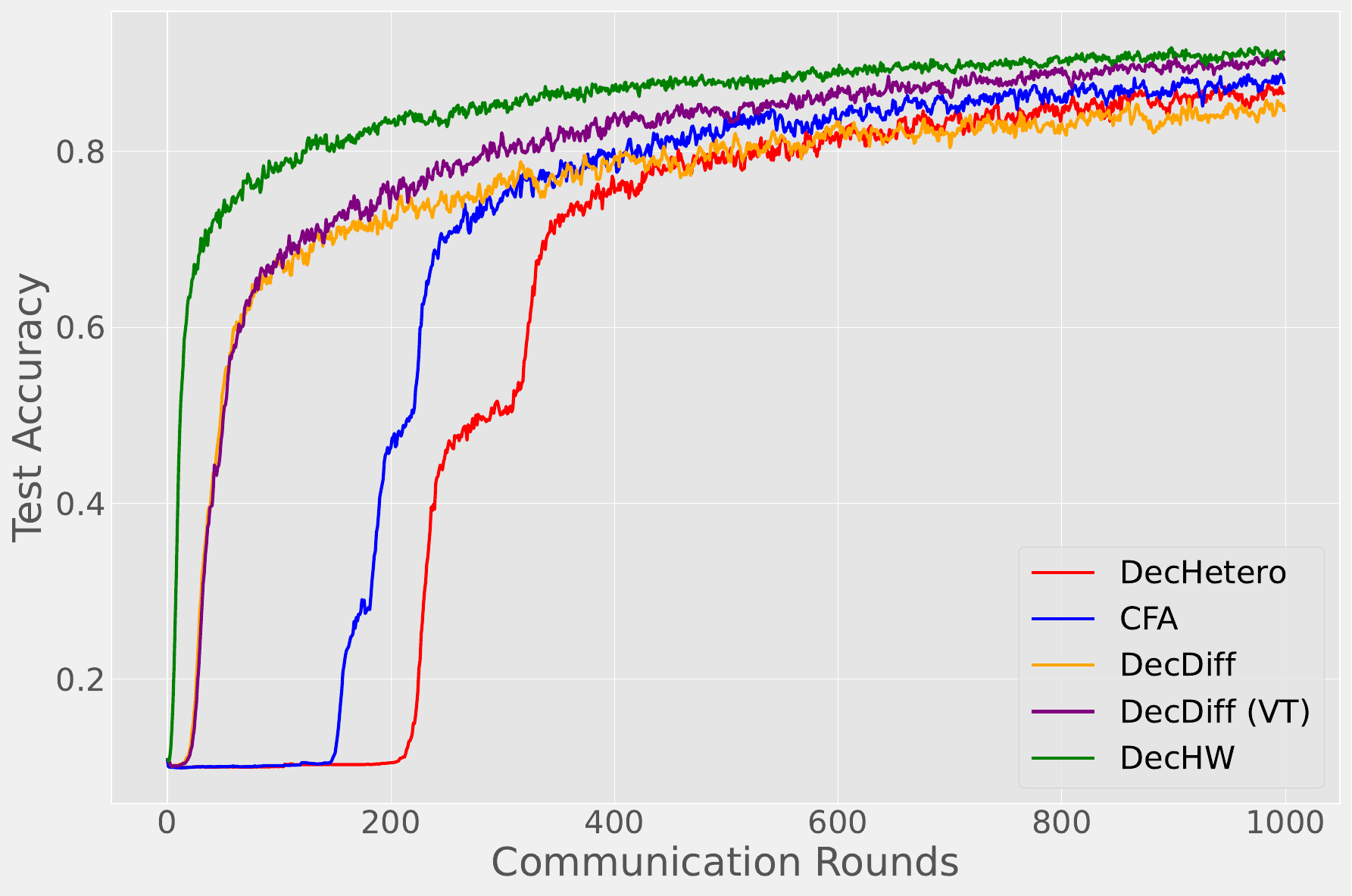}
        \label{fig:mnist_dir05_acc}
    }
    \hfill
    \subfloat[Accuracy (Dir(0.2))]{%
        \includegraphics[width=0.32\textwidth]{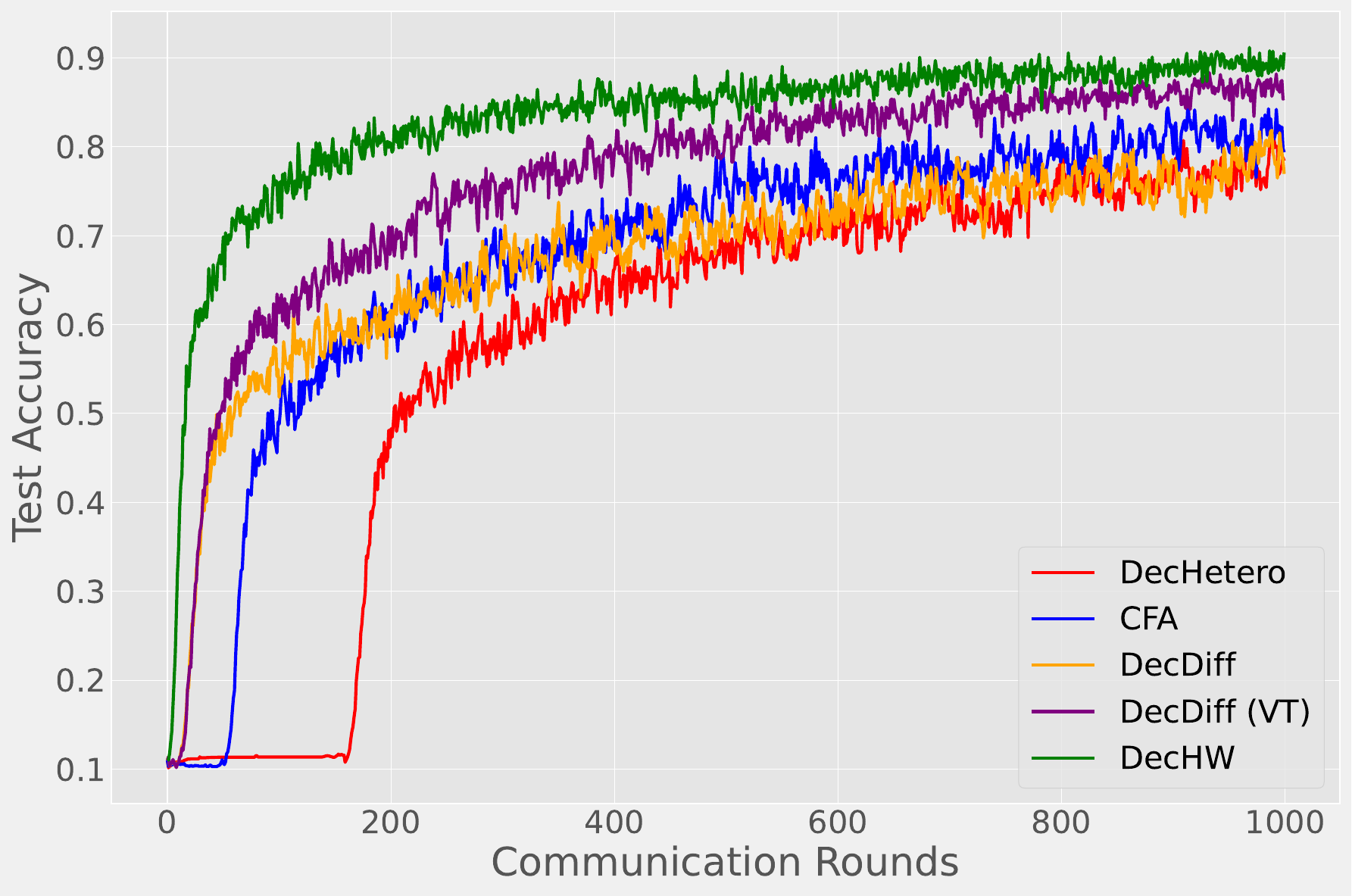}
        \label{fig:mnist_dir02_acc}
    }

    \subfloat[Loss (Dir(1))]{%
        \includegraphics[width=0.32\textwidth]{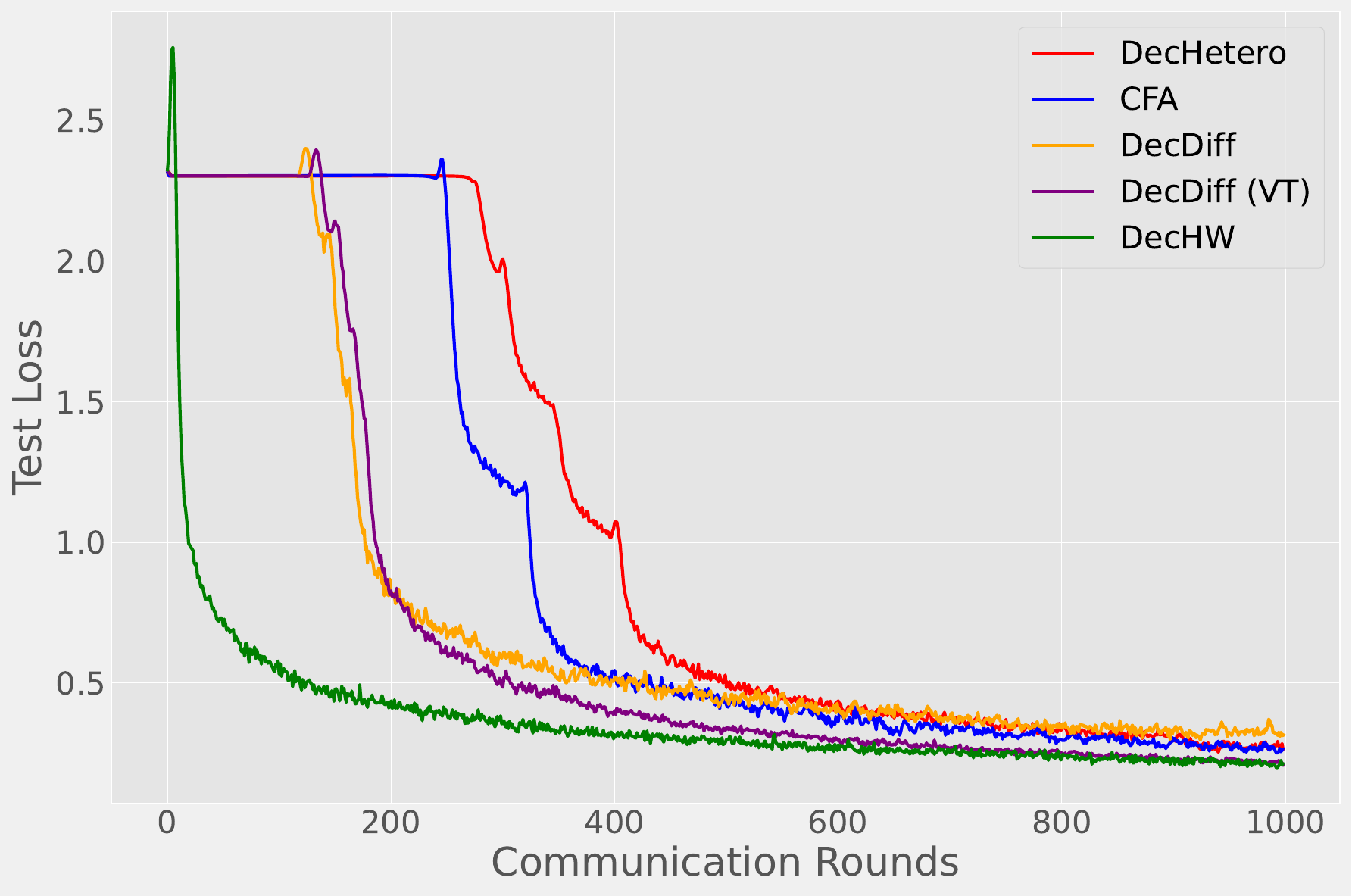}
        \label{fig:mnist_dir1_loss}
    }
    \hfill
    \subfloat[Loss (Dir(0.5))]{%
        \includegraphics[width=0.32\textwidth]{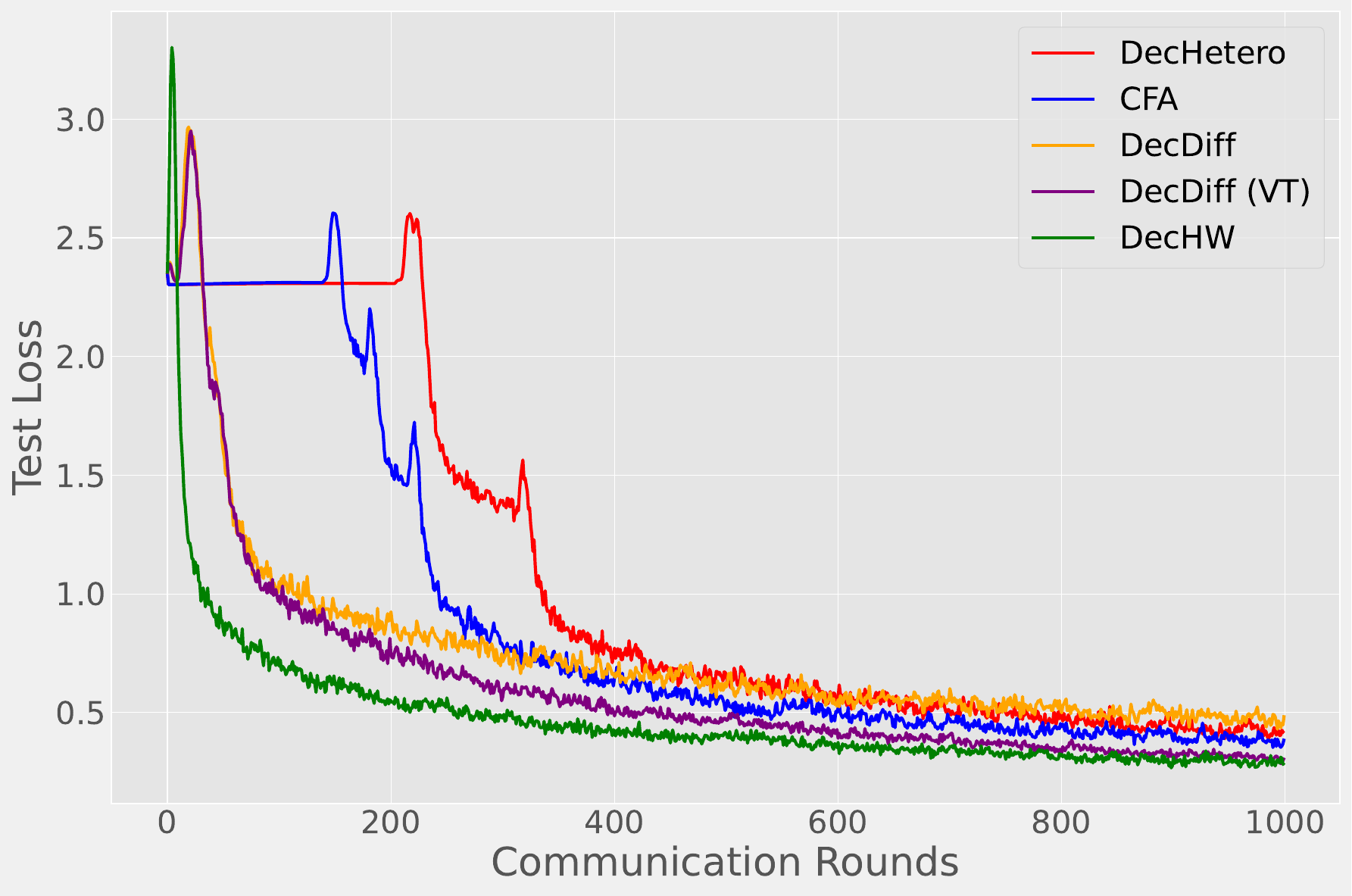}
        \label{fig:mnist_dir05_loss}
    }
    \hfill
    \subfloat[Loss (Dir(0.2))]{%
        \includegraphics[width=0.32\textwidth]{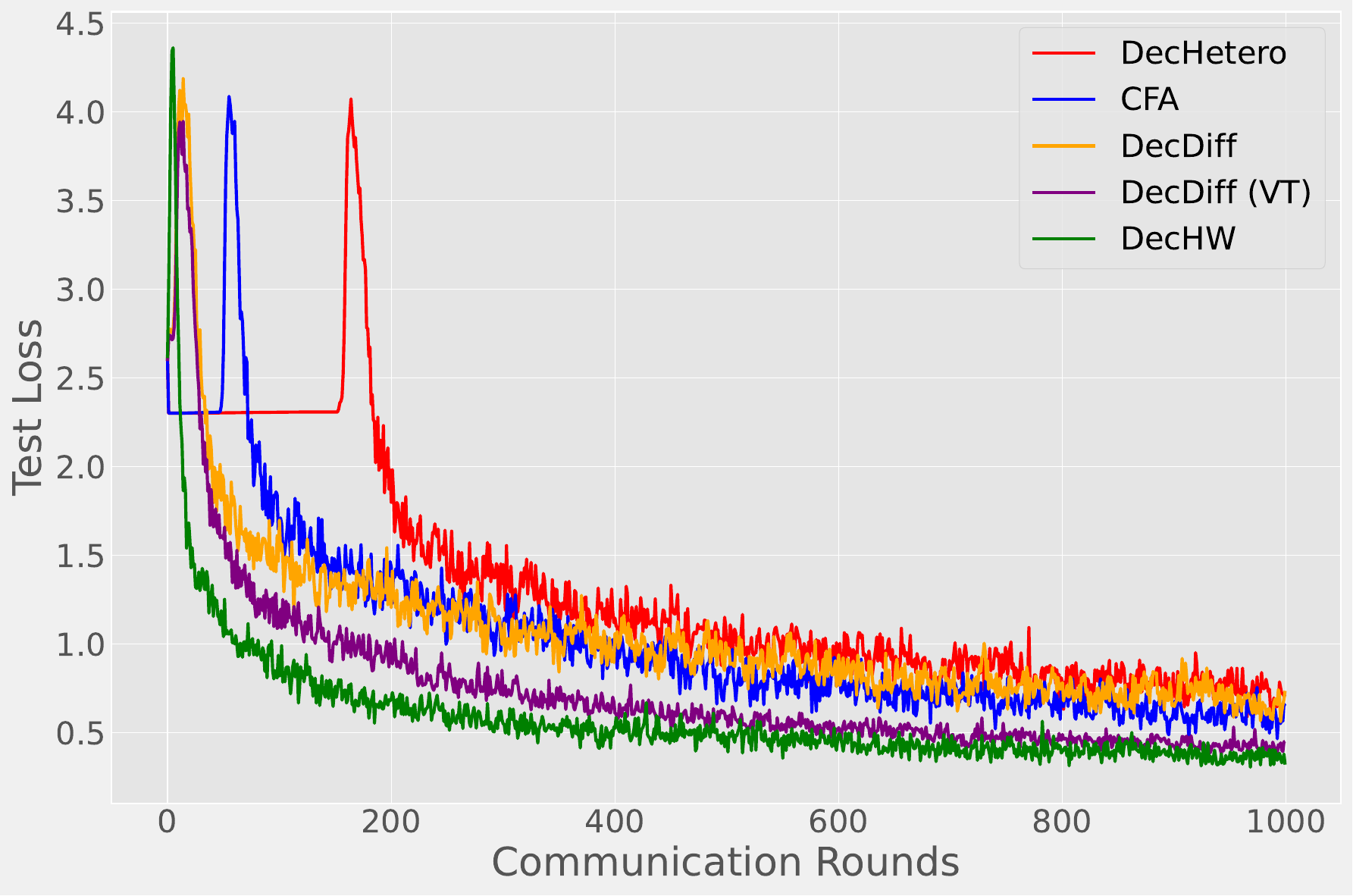}
        \label{fig:mnist_dir02_loss}
    }

    \caption{Average accuracy and loss across devices on the MNIST dataset under various Dirichlet ($\text{Dir}(\alpha)$) distributions. Mean values from 5 runs are plotted over communication rounds.}
    \label{fig:mnist_DFL}
\end{figure*}
\begin{table*}[t]
  \centering
    \caption{ Average time (measured in communication rounds) required to achieve a certain level of accuracy. Empty cells indicate that the method did not achieve the specified accuracy within 1000 rounds.}
  \resizebox{\textwidth}{!}{%
    \begin{tabular}{cccccccccccccccc}
      \toprule
      Dir ($\alpha$) & Method & \multicolumn{4}{c}{MNIST} & \phantom{a} & \multicolumn{4}{c}{Fashion} & \phantom{a} & \multicolumn{4}{c}{CIFAR10} \\
      \cmidrule{3-6} \cmidrule{8-11} \cmidrule{13-16}
      & & $50 \%$ & $75 \%$ & $85 \%$ & $90 \%$ && $50 \%$ & $70 \%$ & $75 \%$ & $80 \%$ && $25 \%$ & $35 \%$ & $45 \%$ & $50 \%$  \\ \midrule
       \multirow{6}{*}{$\alpha = 1$} & DecHetero      & 349 & 412 & 548 & 841 && 8 & 57 & 138 & 448 && 68 & 177 & 400 & 667\\
            & CFA              & 269 & 331 & 460 & 761 && 6 & 39 & 111 & 358 && 48 & 154 & 313 & 513\\
            & DecDiff          & 164 & 212 & 436 & 872 && 8 & 34 & 118 & 395 && 59 & 170 & 368 & 700\\
            & DecDiff (VT)     & 175 & 211 & 331 & 538 && 8 & 39 & 92 & 326 && 72 & 173 & 401 & 703\\
            & DecHW            & \textbf{11} & \textbf{34} &\textbf{ 120} & \textbf{327} && \textbf{5} & \textbf{22} & \textbf{41} & \textbf{259 }&& \textbf{37} & \textbf{83} & \textbf{203 }& \textbf{310}\\
       \midrule

       \multirow{6}{*}{$\alpha = 0.5$} & DecHetero       & 276 & 380 & 740 & - && 7 & 49 & 150 & 522 && 97 & 160 & 311 & 469\\
            & CFA              & 219 & 301 & 610 & - && \textbf{6 }& 44 & 113 & 340 && 100 & 159 & 284 & 452\\
            & DecDiff          & 48 & 245 & 859 & - && 10 & 47 & 104 & 407 && 100 & 160 & 347 & 537\\
            & DecDiff (VT)     & 51 & 191 & 456 & 848 && 10 & 38 & 106 & 368 && 94 & 152 & 325 & 468\\
            & DecHW            &\textbf{ 12} &\textbf{ 64} &\textbf{ 259} & \textbf{638 }&& 8 & \textbf{22} & \textbf{60} & \textbf{288} &&\textbf{ 38} & \textbf{69 }&\textbf{ 130} & \textbf{204}\\
       \midrule
       \multirow{6}{*}{$\alpha = 0.2$} & DecHetero        & 203 & 649 & - & - && 16 & 188 & 525 & - && 199 & 439 & 697 & 977\\
            & CFA              & 90 & 458 & 913 & - && 13 & 142 & 473 & - && 164 & 300 & 456 & 765\\
            & DecDiff          & 55 & 519 & - & - && 17 & 138 & 355 & - && 138 & 238 & 484 & 728\\
            & DecDiff (VT)     & 47 & 237 & 544 & - && 16 & 81 & \textbf{267} & - && 161 & 257 & 429 & 595\\
            & DecHW            & \textbf{17} & \textbf{87} & \textbf{291} & \textbf{673} && \textbf{12} & \textbf{60} & 286 & \textbf{488} && \textbf{56} & \textbf{108} & \textbf{263} & \textbf{447}\\
       \bottomrule

    \end{tabular}%
  }
  \label{tab:communication_effic_tab1}
\end{table*}

\section{Additional Results} \label{app:further_exp}

\begin{figure*}[b]
    \centering

    \subfloat[Accuracy (Dir(1))]{%
        \includegraphics[width=0.3\textwidth]{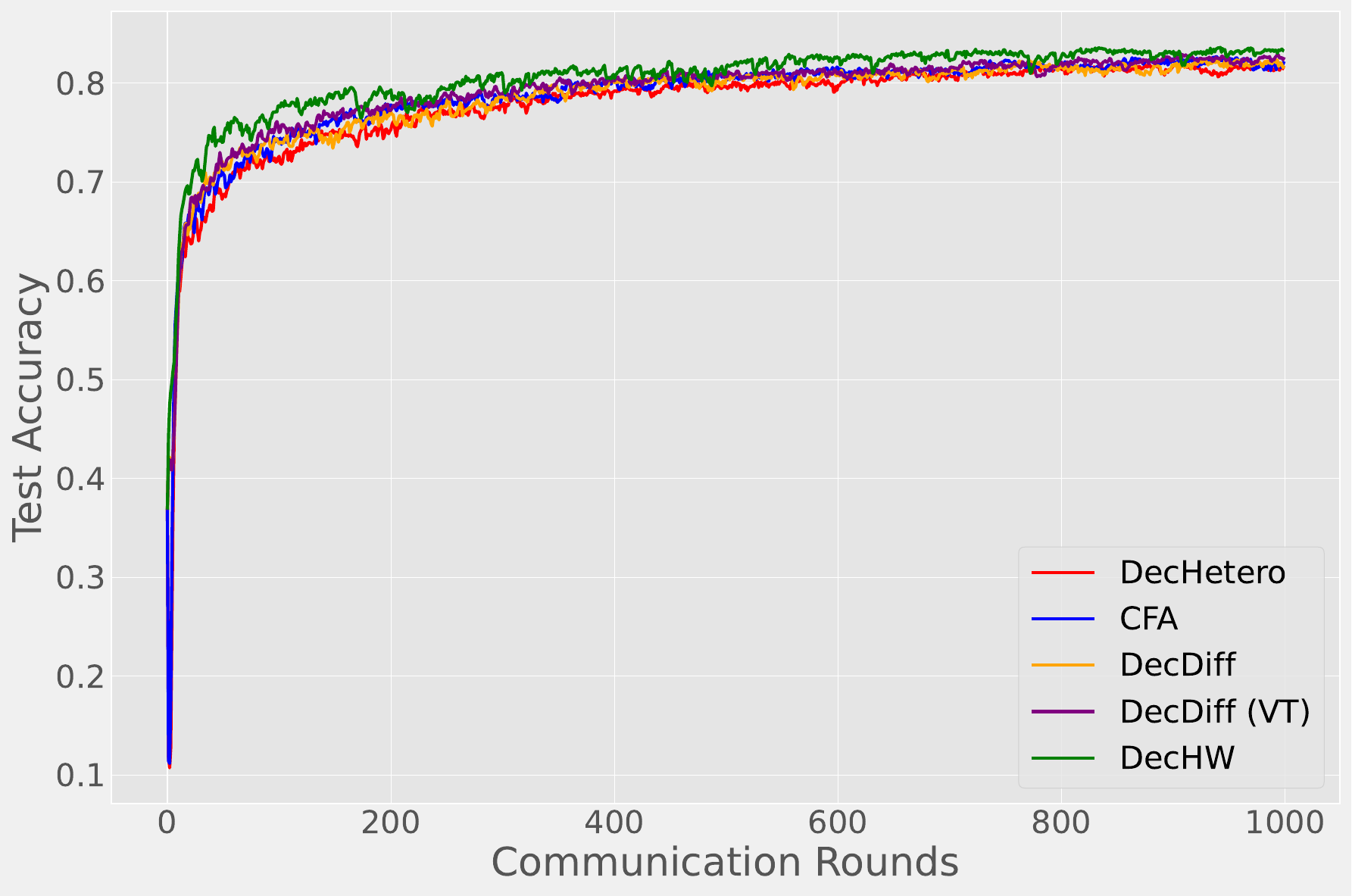}
        \label{fig:fdir1_acc}
    }
    \hfill
    \subfloat[Accuracy (Dir(0.5))]{%
        \includegraphics[width=0.3\textwidth]{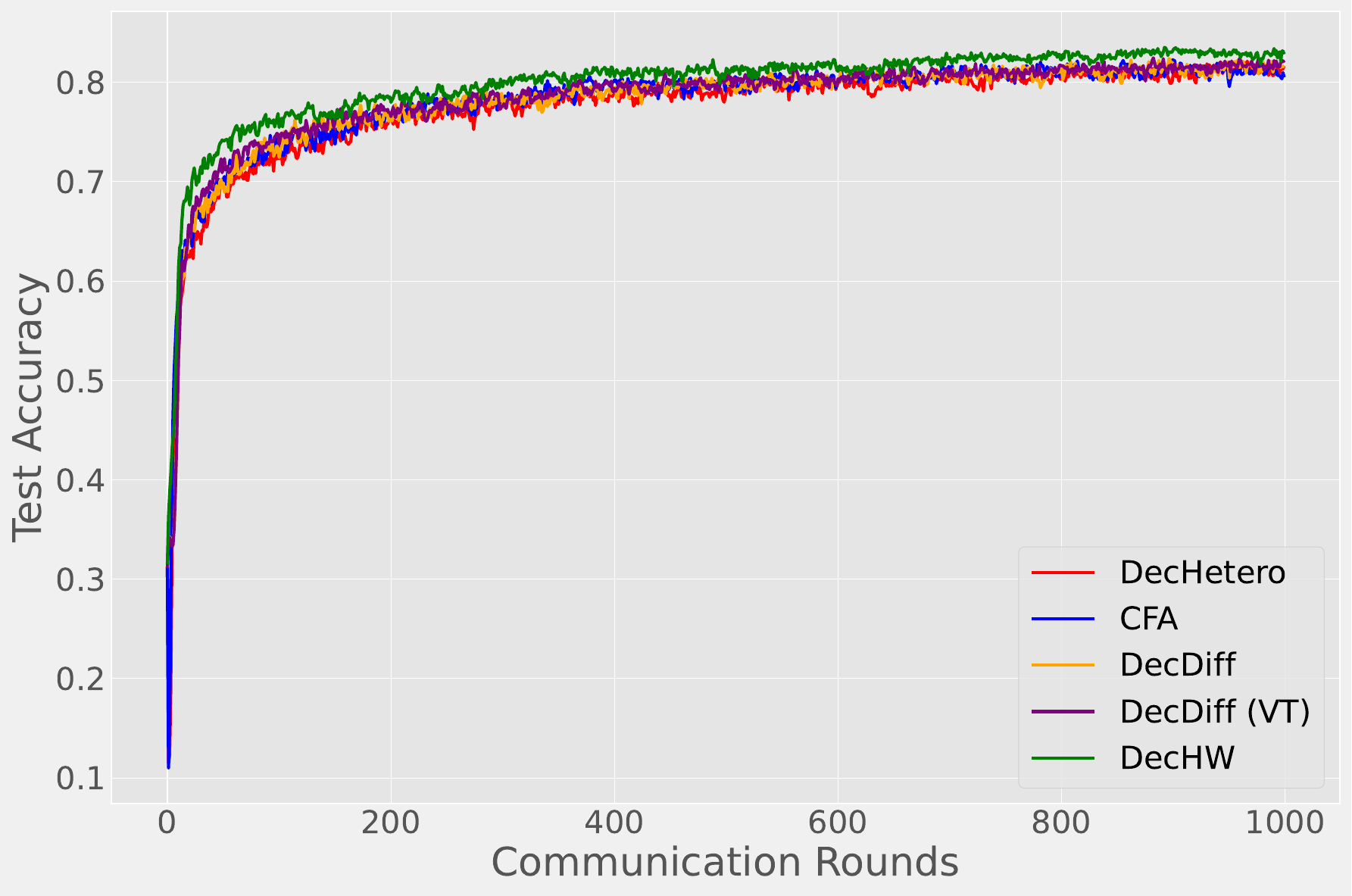}
        \label{fig:fdir05_acc}
    }
    \hfill
    \subfloat[Accuracy (Dir(0.2))]{%
        \includegraphics[width=0.3\textwidth]{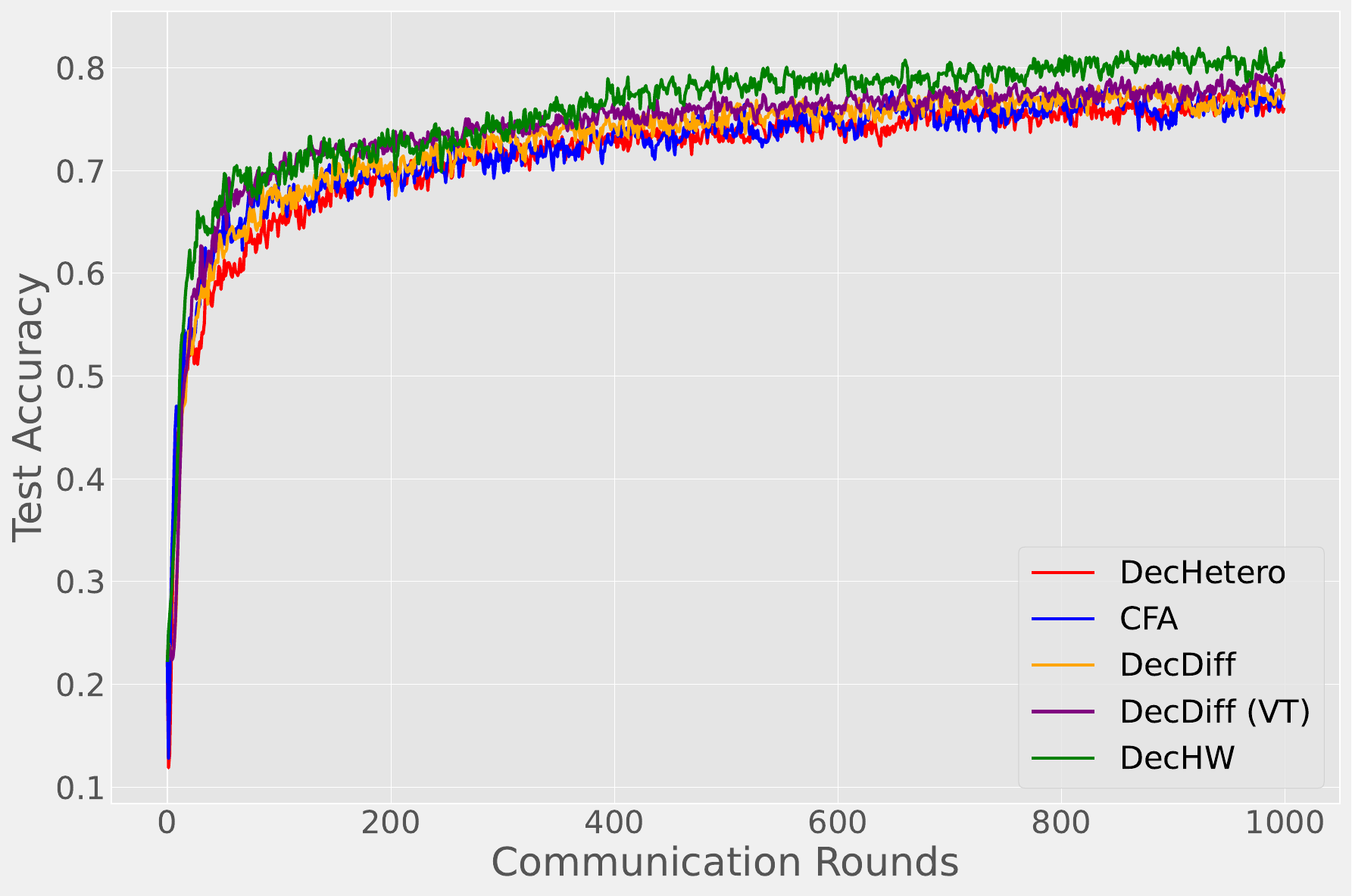}
        \label{fig:fdir02_acc}
    }

    \vspace{1em}

    \subfloat[Loss (Dir(1))]{%
        \includegraphics[width=0.3\textwidth]{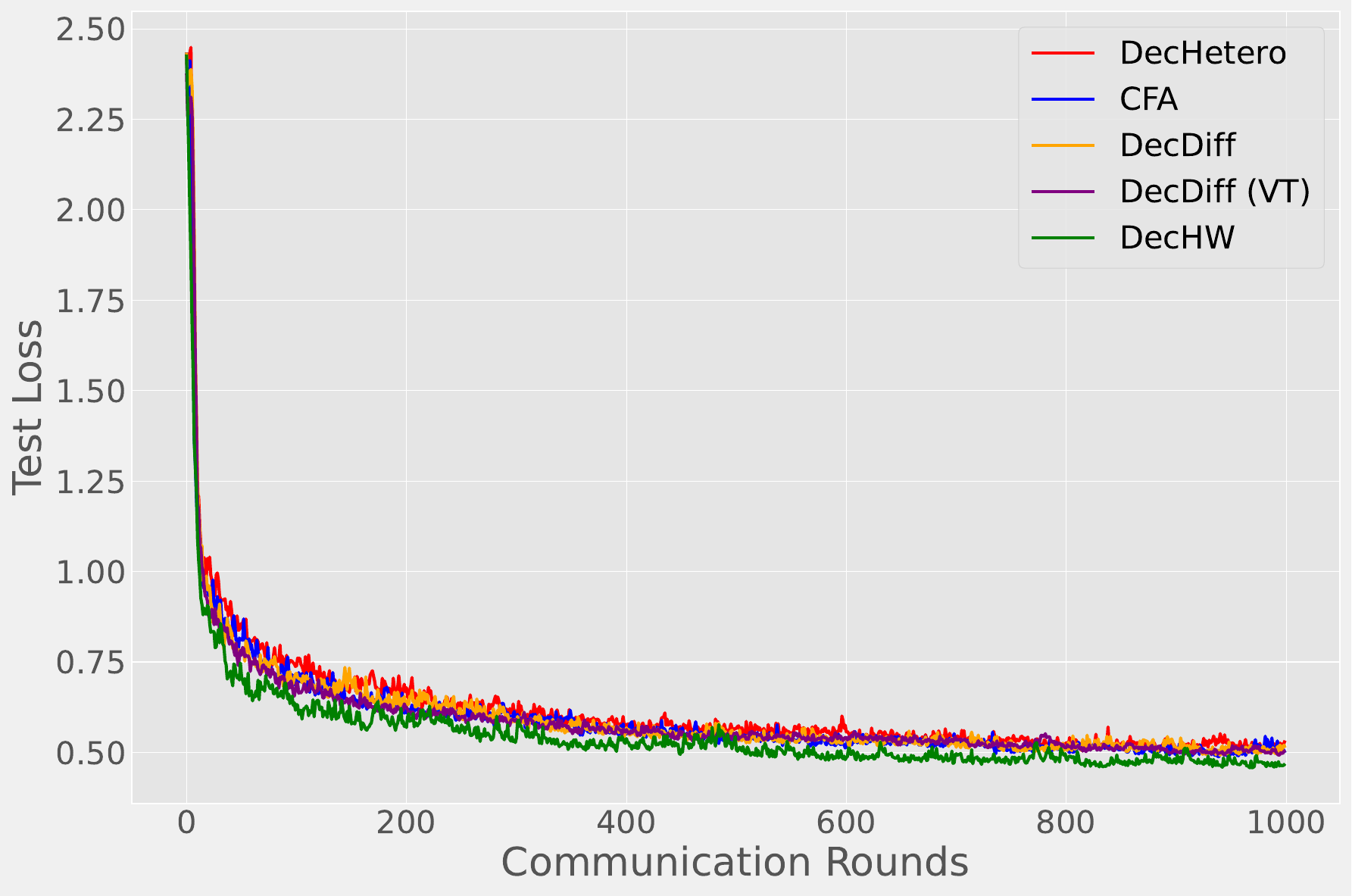}
        \label{fig:fdir1}
    }
    \hfill
    \subfloat[Loss (Dir(0.5))]{%
        \includegraphics[width=0.3\textwidth]{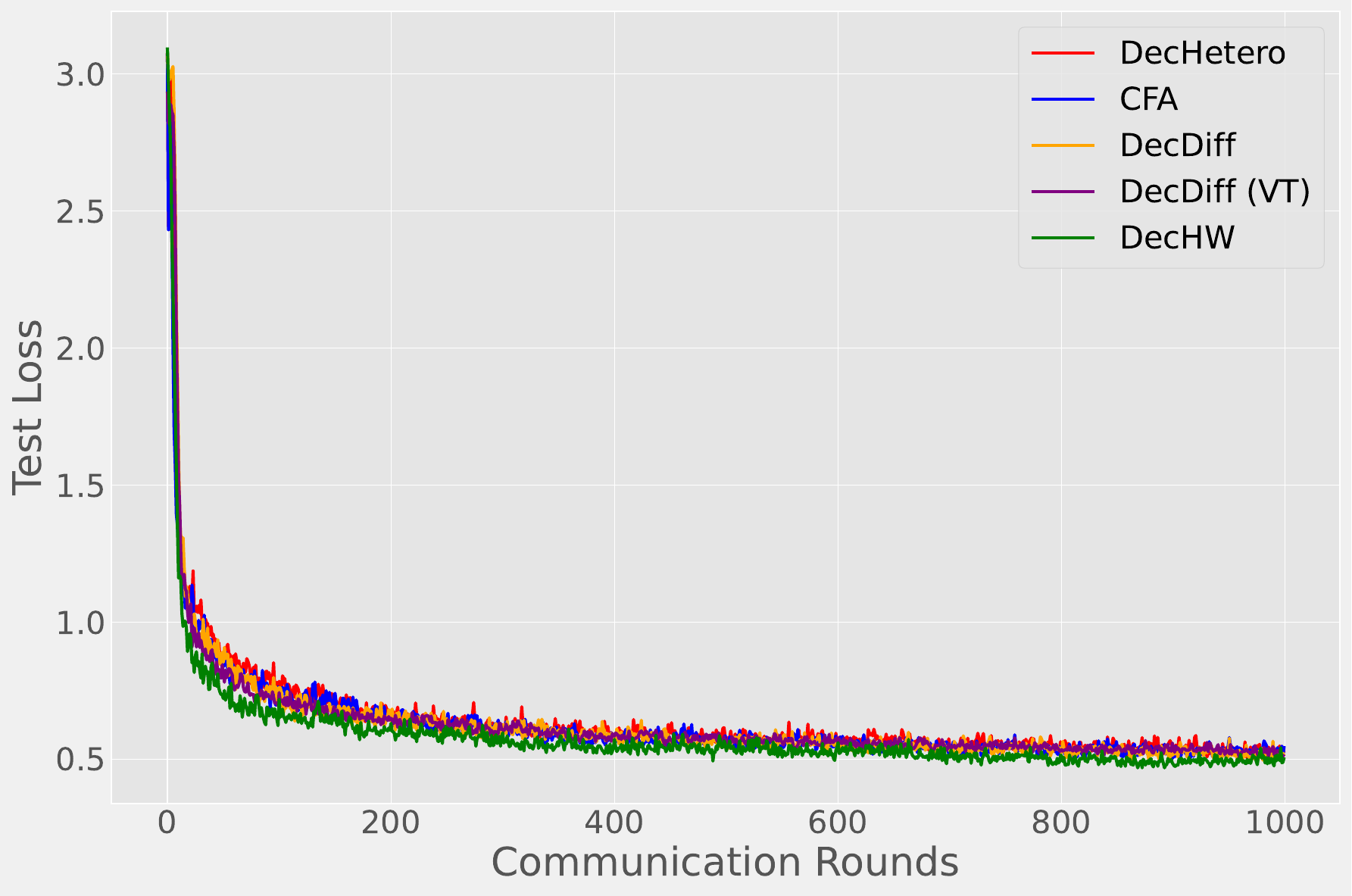}
        \label{fig:fdir0.5}
    }
    \hfill
    \subfloat[Loss (Dir(0.2))]{%
        \includegraphics[width=0.3\textwidth]{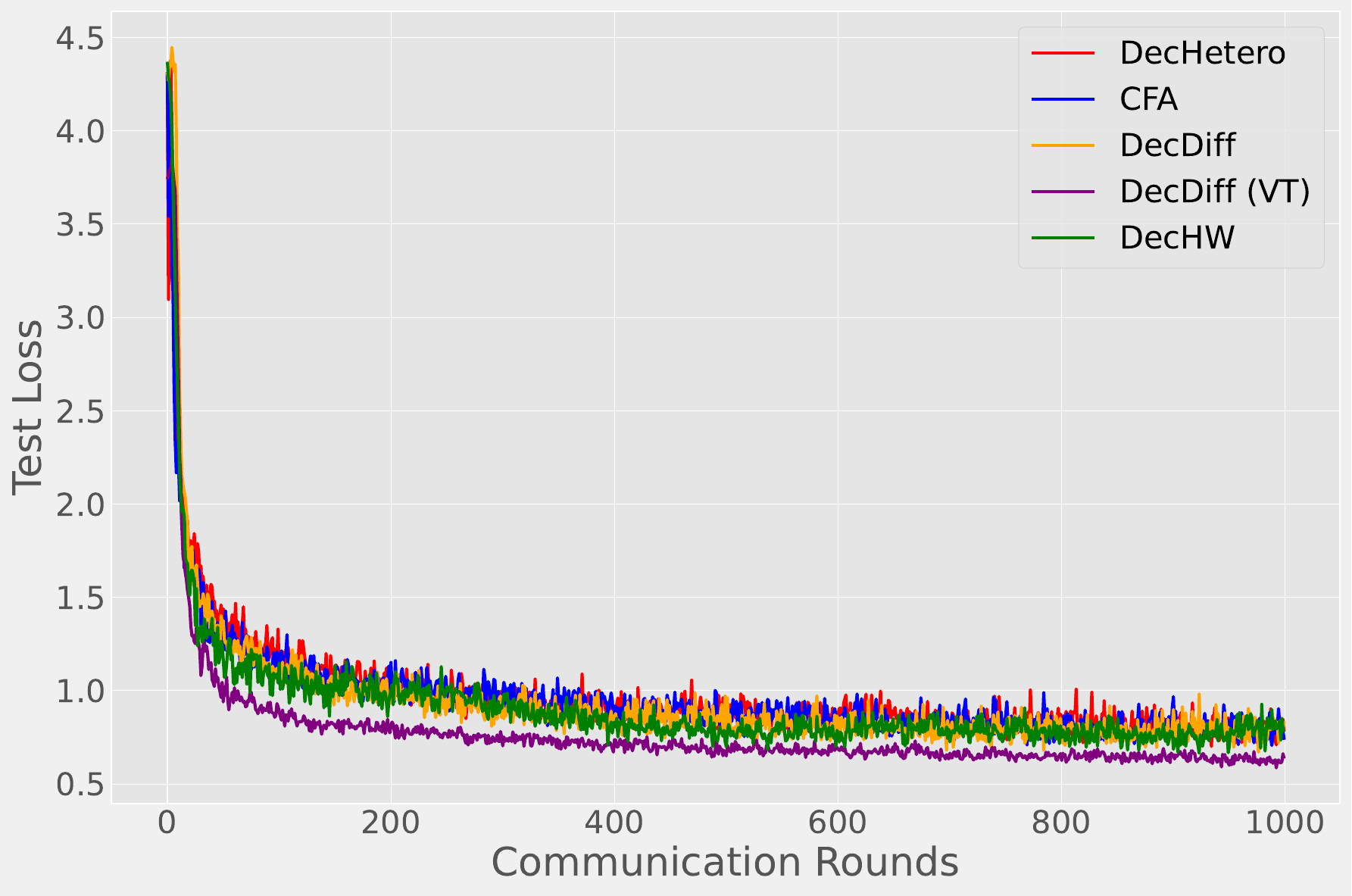}
        \label{fig:fdir0.2}
    }

    \caption{Average accuracy and loss across devices on the Fashion-MNIST dataset under various Dirichlet ($\text{Dir}(\alpha)$) distributions. Mean values from 5 different runs are reported as a function of communication rounds.}
    \label{fig:fmnist_DFL}
\end{figure*}

Here we report the results for MNIST and Fashion-MNIST that were not provided in the body of the paper.
Figure~\ref{fig:mnist_DFL} shows the training curves for the MNIST dataset. It is evident that data heterogeneity and the differences in model initialization significantly affect the performance of DecHetero and CFA in the initial rounds of communication, as they lack strategies to mitigate these challenges. In contrast, DecDiff and DecDiff (VT) struggle initially but gradually improve as training progresses. The proposed method, DecHW, successfully deals with these heterogeneities, achieving fast convergence from the early rounds and maintaining consistent performance throughout the training process. Similarly, as shown in Figure~\ref{fig:fmnist_DFL}, DecHW consistently outperforms the baseline methods on the Fashion-MNIST dataset.

\begin{table*}[t]
  \centering
    \caption{ Average communication rounds required to reach 50\%, 75\%, 90\%, and 95\% of the maximum achievable accuracy for each method across different datasets.}
  \resizebox{\textwidth}{!}{%
    \begin{tabular}{cccccccccccccccc}
      \toprule
      Dir ($\alpha$) & Method & \multicolumn{4}{c}{MNIST} & \phantom{a} & \multicolumn{4}{c}{Fashion} & \phantom{a} & \multicolumn{4}{c}{CIFAR10} \\
      \cmidrule{3-6} \cmidrule{8-11} \cmidrule{13-16}
      & & $50 \%$ & $75 \%$ & $90 \%$ & $95 \%$ && $50 \%$ & $75 \%$ & $90 \%$ & $95 \%$ && $50 \%$ & $75 \%$ & $90 \%$ & $95 \%$  \\ \midrule
       \multirow{6}{*}{$\alpha = 1$} & DecHetero        & 337 & 406 & 476 & 627 && 6 & 13 & 120 & 306 && 82 & 254 & 577 & 754 \\
            & CFA              & 262 & 325 & 376 & 528 && 5 & 13 & 95 & 252 && 73 & 211 & 434 & 619\\
            & DecDiff          & 153 & 177 & 292 & 490 && \textbf{1} & 12 & 88 & 251 && 63 & 215 & 456  & 618\\
            & DecDiff (VT)     & 172 & 192 & 312 & 458 && 2 & 12 & 81 & 247 && 74 & 221 & 457 & \textbf{575}\\
            & DecHW            & \textbf{10} & \textbf{21 }& \textbf{108 }& \textbf{267}   && \textbf{1} & \textbf{10} & \textbf{41} &\textbf{ 159} && \textbf{53} & \textbf{124} &\textbf{ 337} & 591\\
       \midrule

       \multirow{6}{*}{$\alpha = 0.5$} & DecHetero        & 244 & 330 & 451 & 654 && 5 & 16 & 121 & 280 && 117 & 262 & 465 & 650 \\
            & CFA              & 194 & 236 & 386 & 507 && 5 & 13 & 92 & 216 && 124 & 217 & 393 & 646\\
            & DecDiff          & 41 & 76 & 298 & 508   && 8 & 16 & 87 & 244 && 106 & 237 & 422 & 563\\
            & DecDiff (VT)     & 48 & 101 & 299 & 573  && 8 & 14 & 92 & 247 && 103 & 239 & 392 & 533\\
            & DecHW            & \textbf{11} & \textbf{29} & \textbf{163} & \textbf{333 }  && \textbf{4 }& \textbf{11} & \textbf{59 }& \textbf{172} && \textbf{44 }&\textbf{ 108} & \textbf{264} & \textbf{334}\\
       \midrule
       \multirow{6}{*}{$\alpha = 0.2$} & DecHetero        & 186 & 316 & 573 & 722 && 7 & 36 & 188 & 388 && 170 & 339 & 562 & 709 \\
            & CFA              & 934 & 958 & 973 & 986 && \textbf{6} & 30 & 145 & 377 && 140 & 249 & 430 & 608\\
            & DecDiff          & 34 & 142 & 450 & 581 && 11 & 34 & 149 & 355 && 155 & 323 & 547 & 728\\
            & DecDiff (VT)     & 33 & 110 & 290 & 475 && 11 & 30 & \textbf{105} & \textbf{350} && 184 & 355 & 545 & 699\\
            & DecHW            & \textbf{14 }& \textbf{46} & \textbf{148 }& \textbf{291} && 9 &\textbf{ 19} & 127 & 343 && \textbf{59} &\textbf{ 150} &\textbf{ 303} & \textbf{517}\\
       \bottomrule

    \end{tabular}%
  }
  \label{tab:communication_effic_tab2}
\end{table*}
Table \ref{tab:communication_effic_tab1} presents the total number of communication rounds required by each method to achieve a certain level of accuracy. The proposed approach, DecHW, consistently outperforms the baseline by reaching a certain level of absolute accuracy in fewer rounds. This highlights the fast convergence of the proposed approach, even in the presence of data heterogeneity and variations in local model initializations. Furthermore, this performance remains stable throughout the training process, with DecHW achieving subsequent accuracy milestones in fewer rounds and maintaining its position as a top performer in terms of convergence speed. Moreover, Table \ref{tab:communication_effic_tab2} presents the number of rounds each approach requires to achieve a certain level of its own maximum achievable accuracy (reported in Table 1), therefore showcasing how fast each approach is in reaching their best performance. The superior performance of the proposed approach remains consistent across varying degrees of data distribution.

\bibliographystyle{IEEEtran}
\bibliography{ref}
\end{document}